\documentclass[10pt,journal,compsoc]{IEEEtran}

\makeatletter
\def\endthebibliography{%
	\def\@noitemerr{\@latex@warning{Empty `thebibliography' environment}}%
	\endlist
}
\makeatother

\ifCLASSOPTIONcompsoc
  \usepackage[nocompress]{cite}
\else
  \usepackage{cite}
\fi

\usepackage{hyperref}       
\hypersetup{colorlinks=true,urlcolor=black}
\usepackage{url}            
\usepackage{booktabs}       
\usepackage{amsfonts}       
\usepackage{nicefrac}       
\usepackage{microtype}      
\usepackage{algorithm}
\usepackage{algorithmic}
\usepackage{graphicx}
\usepackage{epsfig}
\usepackage{amsmath}
\usepackage{amssymb}
\usepackage{amsthm}
\usepackage{subfigure}
\usepackage{booktabs} 
\usepackage{multirow}
\usepackage{etoolbox}
\usepackage{bbm}
\usepackage{sail}
\usepackage{color}
\usepackage{footnote}
\usepackage{enumitem}
\usepackage{times}
\usepackage{ragged2e}
\usepackage{threeparttable}
\usepackage[export]{adjustbox}
\usepackage[table]{xcolor}
\usepackage{arydshln}
\usepackage{amsthm}

\begin{document}

\title{EPMF: Efficient Perception-aware Multi-sensor Fusion for 3D Semantic Segmentation}

\author{
	Mingkui Tan$^{*\dagger}$, Zhuangwei Zhuang$^*$, Sitao Chen, Rong Li, Kui Jia, Qicheng Wang, Yuanqing Li
	\IEEEcompsocitemizethanks{
	    \IEEEcompsocthanksitem Mingkui Tan, Zhuangwei Zhuang, Sitao Chen, and Rong Li are with the School of Software Engineering, South China University of Technology. Mingkui Tan and Zhuangwei Zhuang are also with the Pazhou Laboratory, Guangzhou, China. E-mail: mingkuitan@scut.edu.cn, \{z.zhuangwei, mechenst, selirong\}@mail.scut.edu.cn. 
        \IEEEcompsocthanksitem Kui Jia is with the School of Electronic and Information Engineering, South China University of Technology. E-mail: kuijia@scut.edu.cn.
        \IEEEcompsocthanksitem Qicheng Wang is with Department of Mathematics at the Hong Kong University of Science and Technology and is also with Minieye, Shenzhen, Guangdong, China. E-mail: wangqicheng@minieye.cc.
        \IEEEcompsocthanksitem Yuanqing Li is with the Pazhou Laboratory, Guangzhou, China. E-mail: auyqli@scut.edu.cn.
		\IEEEcompsocthanksitem $^*$Authors contributed equally. $^\dagger$ Corresponding author.
	}
}

\IEEEtitleabstractindextext{
\begin{abstract}
\justifying
We study multi-sensor fusion for 3D semantic segmentation that is important to scene understanding for many applications, such as autonomous driving and robotics. For example, for autonomous cars equipped with RGB cameras and LiDAR, it is crucial to fuse complementary information from different sensors for robust and accurate segmentation. Existing fusion-based methods, however, may not achieve promising performance due to the vast difference between the two modalities. In this work, we investigate a collaborative fusion scheme called perception-aware multi-sensor fusion (PMF) to effectively exploit perceptual information from two modalities, namely, appearance information from RGB images and spatio-depth information from point clouds. To this end, we first project point clouds to the camera coordinate using perspective projection. In this way, we can process both inputs from LiDAR and cameras in 2D space while preventing the information loss of RGB images. Then, we propose a two-stream network that consists of a LiDAR stream and a camera stream to extract features from the two modalities, separately. The extracted features are fused by effective residual-based fusion modules. Moreover, we introduce additional perception-aware losses to measure the perceptual difference between the two modalities. 
Last, we propose an improved version of PMF,~\ie, EPMF, which is more efficient and effective by optimizing data pre-processing and network architecture under perspective projection. Specifically, we propose cross-modal alignment and cropping to obtain tight inputs and reduce unnecessary computational costs. We then explore more efficient contextual modules under perspective projection and fuse the LiDAR features into the camera stream to boost the performance of the two-stream network. Extensive experiments on benchmark data sets show the superiority of our method. For example, on nuScenes test set, our EPMF outperforms the state-of-the-art method,~\ie, RangeFormer, by \textbf{0.9\%} in mIoU. Compared to PMF, EPMF also achieves \textbf{2.06}$\times$ acceleration with \textbf{2.0\%} improvement in mIoU. Our source code is available at \url{https://github.com/ICEORY/PMF}.
\end{abstract}

}

\maketitle

\IEEEdisplaynontitleabstractindextext

%
\IEEEpeerreviewmaketitle

\IEEEraisesectionheading{\section{Introduction}\label{sec:introduction}}
\IEEEPARstart{S}{emantic} scene understanding is a fundamental task for many applications, such as autonomous driving and robotics~\cite{gan2019self,Liu2019LPDNet3P,Rusu2008Towards3P,Shan2018LeGOLOAMLA}. Specifically, in the scenes of autonomous driving, it provides fine-grained environmental information for high-level motion planning and improves the safety of autonomous cars~\cite{behley2019semantickitti,geiger2012we}.
One of the important tasks in semantic scene understanding is semantic segmentation, which assigns a class label to each data point in the input data, and helps autonomous cars to better understand the environment.

\begin{figure}[t]
    \centering
    \includegraphics[width=\columnwidth]{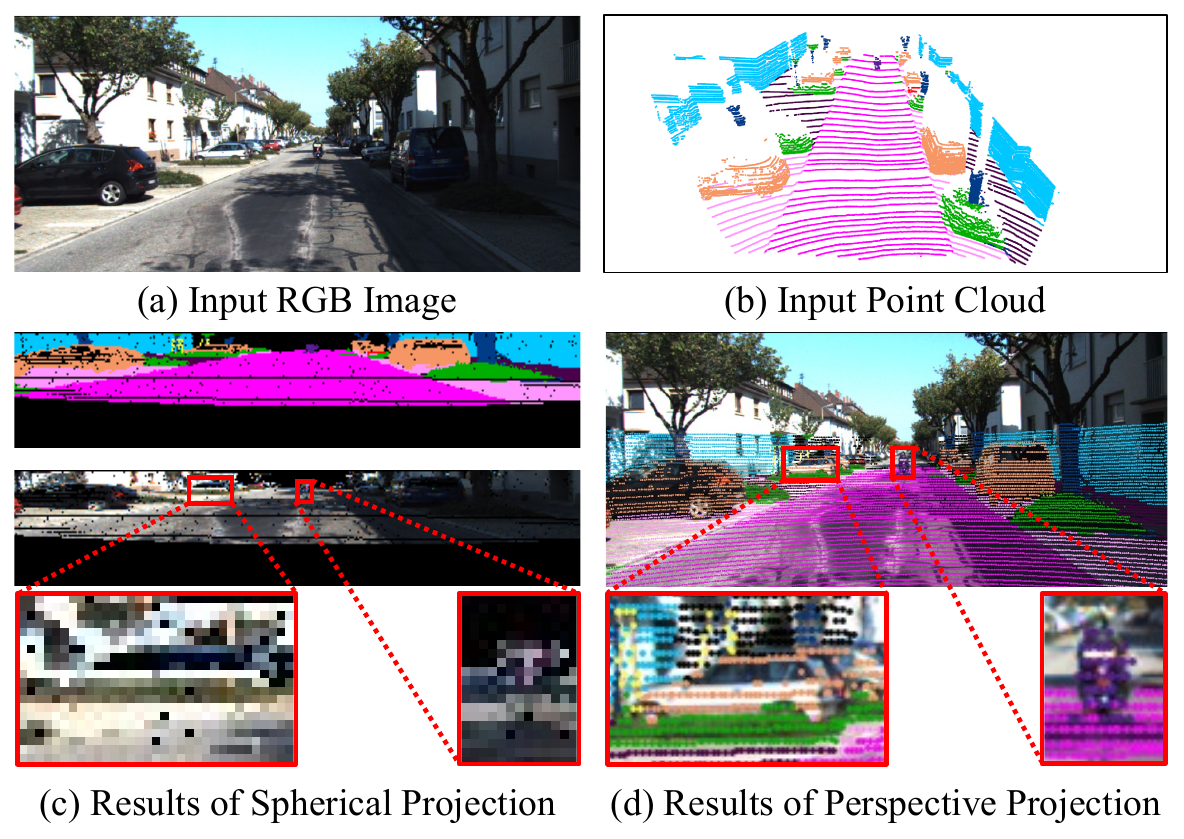}
    \caption{Comparisons of spherical projection~\cite{milioto2019rangenet++,wu2018squeezeseg} and perspective projection. With spherical projection, most of the appearance information from RGB images is lost. Instead, we preserve the information of images with perspective projection. To distinguish different classes, we colorize the point clouds using semantic labels from SemanticKITTI.}
    \label{fig:iv_vs_sv}
\end{figure}

According to the sensors used by semantic segmentation methods, recent studies can be divided into three categories: camera-only methods~\cite{Badrinarayanan2017SegNetAD,chen2017deeplab,chen2017rethinking,Long2015FullyCN,yuan2018ocnet}, LiDAR-only methods~\cite{aksoy2019salsanet,cortinhal2020salsanext,hu2020randla,wu2018squeezeseg,zhang2020polarnet} and multi-sensor fusion methods~\cite{krispel2020fuseseg,Madawy2019RGBAL,meyer2019sensor,vora2020pointpainting,zhang2020deep}. Camera-only methods have achieved great progress with the help of a massive amount of open-access data sets~\cite{brostow2008segmentation,cordts2016cityscapes,dollar2011pedestrian}. Since images obtained by a camera are rich in appearance information (\eg, texture and color), camera-only methods can provide fine-grained and accurate semantic segmentation results. However, as passive sensors, cameras are susceptible to changes in lighting conditions and are thus unreliable~\cite{sitawarin2018darts}.\footnote{See Section~\ref{sec:abla_anti_attack} for more details.}
To address this problem, researchers conduct semantic segmentation on point clouds from LiDAR.
Compared with camera-only approaches, LiDAR-only methods are more robust in different light conditions, as LiDAR provides reliable and accurate spatio-depth information on the physical world.
Unfortunately, LiDAR-only semantic segmentation is challenging due to the sparse and irregular distribution of point clouds. In addition, point clouds lack texture and color information, resulting in high classification error in the fine-grained segmentation task of LiDAR-only methods. 
A straightforward solution for addressing both drawbacks of camera-only and LiDAR-only methods is to fuse the multimodal data from both sensors,~\ie, multi-sensor fusion methods. Nevertheless, due to the large domain gap between RGB cameras and LiDAR, multi-sensor fusion is still a nontrivial task.

In multi-sensor fusion methods, fusing multimodal data from different sensors is an important problem. Existing fusion-based methods~\cite{Madawy2019RGBAL,vora2020pointpainting} mainly lift the dense 2D image features to the 3D LiDAR coordinates using spherical projection~\cite{milioto2019rangenet++} and conduct feature fusion in the sparse LiDAR domain. However, these methods suffer from a critical limitation: as the point clouds are very sparse, most of the appearance information from the RGB images is missing after un-projecting it to the LiDAR coordinates. For example, as shown in Figure~\ref{fig:iv_vs_sv} (c), the car and motorcycle in the image become distorted with spherical projection. As a result, existing fusion-based methods have difficulty capturing the appearance information from the projected RGB images.

\begin{figure}[t]
    \centering
    \includegraphics[width=\linewidth]{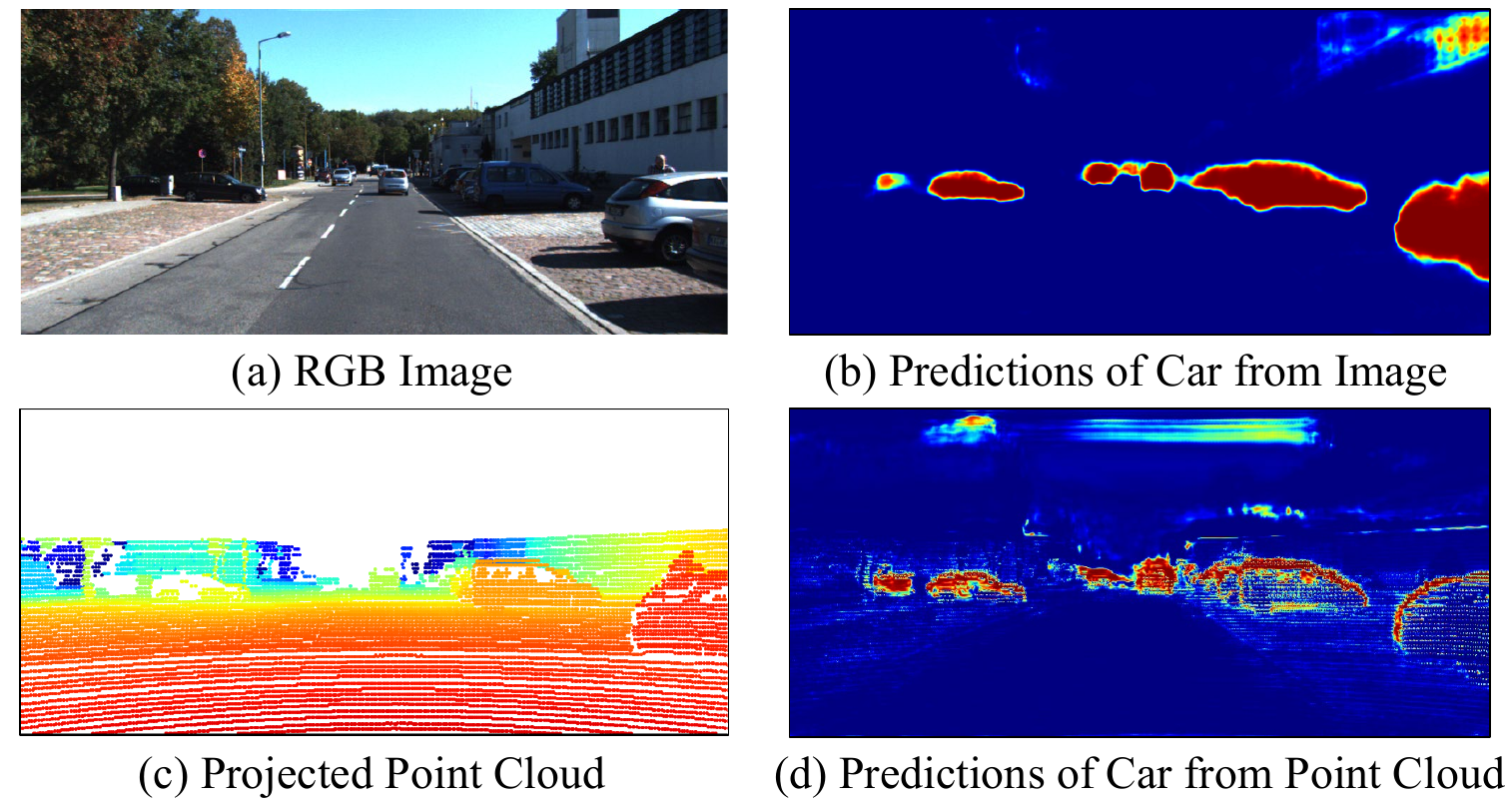}
    \caption{Comparisons of the predictions from images and point clouds. Deep neural networks capture different perceptual information from RGB images and point clouds. Red indicates predictions with higher scores.}
    \label{fig:perceptual_confidence}
    \vskip -0.15in
\end{figure}

In this paper, we aim to exploit an effective multi-sensor fusion method. Unlike existing methods~\cite{Madawy2019RGBAL,vora2020pointpainting}, we assume and highlight that the perceptual information from both RGB images and point clouds,~\ie, appearance information from images and spatio-depth information from point clouds, is important in fusion-based semantic segmentation. Based on this intuition, we propose a perception-aware multi-sensor fusion (PMF) scheme that conducts a collaborative fusion of perceptual information from two modalities of data in three aspects. \textbf{First}, we propose perspective projection to project the point clouds to the camera coordinate system to obtain additional spatio-depth information for RGB images. \textbf{Second}, we propose a two-stream network (TSNet) that contains a camera stream and a LiDAR stream to extract perceptual features from multi-modal sensors separately. Considering that the information from images is unreliable in an outdoor environment, we fuse the image features to the LiDAR stream by effective residual-based fusion (RF) modules, which are designed to learn the complementary features of the original LiDAR modules. \textbf{Third}, we propose perception-aware losses to measure the vast perceptual difference between the two data modalities and boost the fusion of different perceptual information. Specifically, as shown in Figure~\ref{fig:perceptual_confidence}, the perceptual features captured by the camera stream and LiDAR stream are different. Therefore, we use the predictions with higher confidence to supervise those with lower confidence. Since model efficiency is also an essential factor for real-world applications, we further exploit the efficiency of PMF and propose an improved version,~\ie, EPMF.

\begin{figure}
    \centering
    \includegraphics[width=0.95\linewidth]{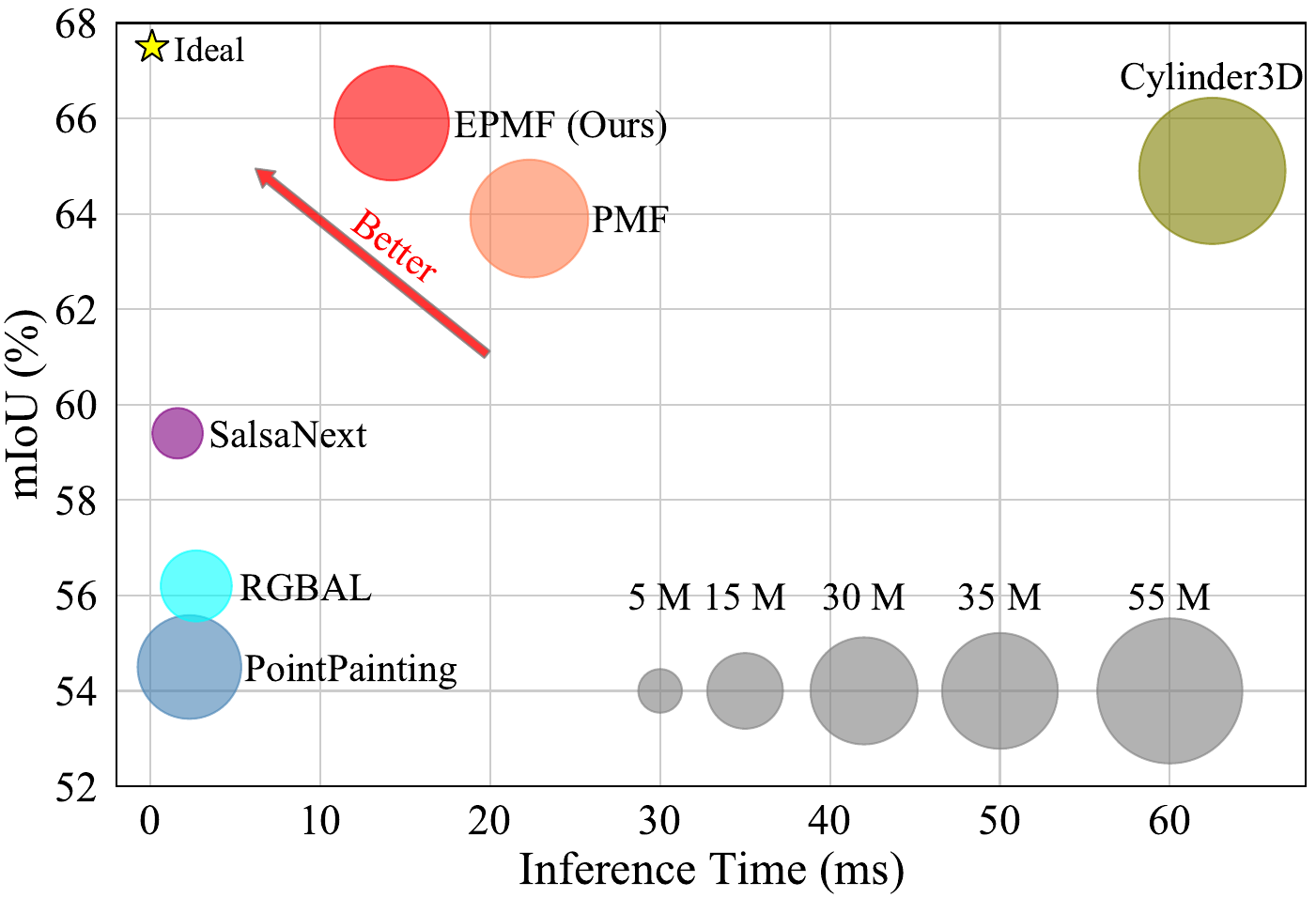}
    \caption{Comparisons of efficiency and performance of different methods on SemanticKITTI-FV.}
    \label{fig:eff_cmp_smk}
\end{figure}

Our contributions are summarized as follows. First, we propose a perception-aware multi-sensor fusion (PMF) scheme to effectively fuse the perceptual information from RGB images and point clouds. Second, by fusing the spatio-depth information from point clouds and appearance information from RGB images, PMF is able to address segmentation with undesired light conditions and sparse point clouds. More critically, PMF is robust in adversarial samples of RGB images by integrating the information from point clouds.
Third, we introduce perception-aware losses into the network and force the network to capture the perceptual information from two different-modality sensors. As demonstrated in Figure~\ref{fig:eff_cmp_smk}, on top of PMF, we further propose EPMF, which reduces the model complexity of PMF while improving the model performance by a large margin. The extensive experiments on three benchmark data sets including SemanticKITTI-FV~\cite{behley2019semantickitti}, nuScenes~\cite{caesar2020nuscenes}, and A2D2~\cite{geyer2020a2d2}, demonstrate the superior performance of our method. For example, 
on nuScenes test set, our EPMF outperforms the state-of-the-art methods,~\ie, SphereFormer~\cite{lai2023spherical} and RangeFormer~\cite{kong2023rethinking} by 1.1\% and 0.9\% in mIoU, respectively, without additional fine-tuning or test-time augmentation.

This paper extends our prior version\cite{zhuang2021perception} from following aspects. 1) We propose cross-modal alignment and cropping (CAC) to address the miss-alignment issue of point clouds and RGB images. 2) We explore the impact of the different resolutions of point clouds and improve the efficiency of our method without performance degradation. 3) We adopt the proposed EPMF on more benchmark data sets and show the superior performance of our method on extremely sparse point clouds. 4) We provide more ablation studies to investigate the effectiveness of our method.

\section{Related Work}
In this section, we revisit the existing literature on 2D and 3D semantic segmentation,~\ie, camera-only methods, LiDAR-only methods and multi-sensor fusion methods. 

\noindent\textbf{Camera-only methods}
Camera-only semantic segmentation aims to predict the pixel-wise labels of 2D images. FCN~\cite{Long2015FullyCN} is a fundamental work in semantic segmentation, which proposes an end-to-end fully convolutional architecture based on image classification networks. In addition to FCN, recent works have achieved significant improvements via exploring multi-scale information~\cite{chen2017deeplab, lin2016efficient, zhao2017pyramid}, dilated convolution~\cite{chen2017rethinking,mehta2018espnet,wang2018understanding}, and attention mechanisms~\cite{huang2019ccnet,yuan2018ocnet}. 
More recently, transformer-based methods have been proposed for robust and accurate segmentation~\cite{xie2021segformer,cheng2022masked}. Specifically, SegFormer~\cite{xie2021segformer} designs a positional-encoding-free and hierarchical transformer encoder that generates both high-resolution fine features and low-resolution coarse features. Moreover, it proposes a lightweight All-MLP decoder to aggregate the multiscale features for robust semantic segmentation. Mask2Former~\cite{cheng2022masked} proposes a universal architecture to address any image segmentation tasks, including panoptic, instance, or semantic segmentation. Although these methods show strong robustness to the corruptions and perturbations in autonomous driving scenes, their performance under poor lighting conditions (\ie, at night-time) is unsatisfactory compared to LiDAR-based methods.

\noindent\textbf{LiDAR-only methods}
To address the drawbacks of cameras, LiDAR is an important sensor on an autonomous car, as it is robust in more complex scenes. 
According to the preprocessing pipeline, existing methods for point clouds mainly contain two categories, including direct methods~\cite{hu2020randla,qi2017pointnet,qi2017pointnet++,zhu2021cylindrical} and projection-based methods~\cite{cortinhal2020salsanext, wu2018squeezeseg, wu2019squeezesegv2, xu2020squeezesegv3}.

Direct methods perform semantic segmentation by processing the raw 3D point clouds directly. PointNet~\cite{qi2017pointnet} is a pioneering work in this category that extracts point cloud features by multi-layer perception. A subsequent extension,~\ie, PointNet++~\cite{qi2017pointnet++}, further aggregates a multi-scale sampling mechanism to aggregate global and local features. However, these methods do not consider the varying sparsity of point clouds in outdoor scenes.
Cylinder3D~\cite{zhu2021cylindrical} addresses this issue by using 3D cylindrical partitions and asymmetrical 3D convolutional networks. SphereFormer~\cite{lai2023spherical} directly aggregates information from dense close points to sparse distant ones through radial window partitions and proposes dynamic feature selection to select local neighbor features or radial contextual features. However, direct methods have a high computational complexity, which limits their applicability in autonomous driving. PVKD~\cite{hou2022point} transfers both point-level and voxel-level hidden knowledge from a large LiDAR semantic segmentation model to a slim network to achieve model compression. WaffleIron~\cite{puy2023using} uses standard MLPs and dense 2D convolutions to build a 3D backbone for point cloud semantic segmentation, which does not rely on sparse 3D convolution. In addition, one can easily improve the efficiency of networks by existing neural architecture search~\cite{cai2019once,guo2021towards,niu2021disturbance} and model compression techniques~\cite{han2015deep,liu2021discrimination,xu2020generative}.

Projection-based methods are more efficient because they convert 3D point clouds to a 2D grid. 
In projection-based methods, researchers focus on exploiting effective projection methods, such as spherical projection~\cite{milioto2019rangenet++,wu2018squeezeseg} and bird's-eye projection~\cite{zhang2020polarnet}. Such 2D representations allow researchers to investigate efficient network architectures based on existing 2D convolutional networks~\cite{aksoy2019salsanet,cortinhal2020salsanext,Guo2019NATNA}.
AMVNet~\cite{liong2020amvnet} utilizes the late fusion to combine the advantages of both the range view and bird's-eye view networks. RPVNet~\cite{xu2021rpvnet} proposes a range-point-voxel fusion network to synergize all the three view’s representations to alleviate the above problem. RangeFormer~\cite{kong2023rethinking} formulates the segmentation of range view grids as a seq2seq problem and adopts several standard Transformer blocks to capture the rich contextual information, then uses the MLP heads to decode the multi-scale features.
Unlike uni-modal methods, we focus on fusing information from both the camera and LiDAR to achieve accurate and robust 3D semantic segmentation for autonomous driving.

\begin{figure*}
    \centering
    \includegraphics[width=1.0\linewidth]{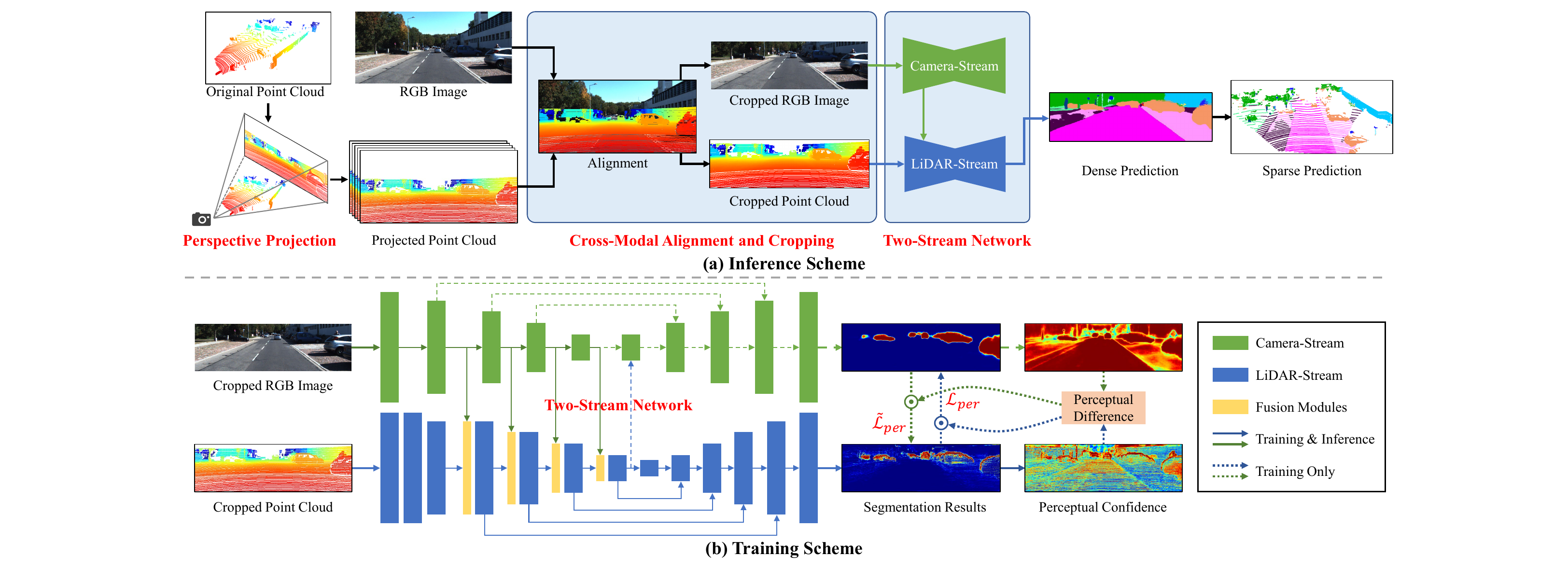}
    \caption{Illustration of the training and inference schemes of EPMF. EPMF consists of three components: (1) perspective projection with cross-modal alignment and crop; (2) a two-stream network (TSNet) with feature fusion modules; and (3) perception-aware losses $\mL_{per},\widetilde{\mL}_{per}$~\wrt~the camera stream and the LiDAR stream. We first project the point clouds to the camera coordinate with perspective projection and learn the features from both the RGB images and point clouds using TSNet. The image features are fused into the LiDAR stream network by fusion modules. In the training procedure, we use perception-aware losses to help the network focus on the perceptual features of both images and point clouds. In the inference procedure, we apply dense-to-sparse mapping to obtain 3D segmentation results of point clouds.}
    \label{fig:arch_overview}
\end{figure*}

\noindent\textbf{Multi-sensor fusion methods}
To leverage the benefits of both camera and LiDAR, recent work has attempted to fuse information from two complementary sensors to improve the accuracy and robustness of the 3D semantic segmentation algorithm~\cite{krispel2020fuseseg,Madawy2019RGBAL,meyer2019sensor,vora2020pointpainting}. RGBAL~\cite{Madawy2019RGBAL} converts RGB images to a polar-grid mapping representation and designs early and mid-level fusion strategies. 
PointPainting~\cite{vora2020pointpainting} obtains the segmentation results of images and projects them to the LiDAR space by using bird's-eye projection~\cite{zhang2020polarnet} or spherical projection~\cite{milioto2019rangenet++}. The projected segmentation scores are concatenated with the original point cloud features to improve the performance of LiDAR networks. 
2DPASS~\cite{yan20222dpass} enhances the representation learning of 3D semantic segmentation network by distilling multi-modal knowledge to single point cloud modality. In this way, 2DPASS can use LiDAR-only input in test-time. However, the model performance is unsatisfactory under the scene with sparser point clouds (\eg, A2D2 with 16-beam LiDARs). In contrast, our EPMF can achieve promising performance by fusing 2D images and 3D point clouds during inference.
Besides, unlike existing methods that perform feature fusion in the LiDAR domain, PMF~\cite{zhuang2021perception} exploits a collaborative fusion of multimodal data in camera coordinates. In this work, we further extend PMF to improve its efficiency and performance.


\section{Proposed Method}

In this work, we propose an efficient perception-aware multi-sensor fusion (EPMF) scheme to perform an effective fusion of the perceptual information from both RGB images and point clouds. Specifically, as shown in Figure~\ref{fig:arch_overview}, EPMF contains three components: (1) perspective projection with cross-modal alignment and cropping; (2) a two-stream network (TSNet) with residual-based fusion modules; (3) perception-aware losses. The general scheme of EPMF is shown in Algorithm~\ref{alg:optimization}. We first project the point clouds to the camera coordinate system by using perspective projection. Then, we use a two-stream network that contains a camera stream and a LiDAR stream to extract perceptual features from the two modalities, separately. The features from the camera stream are fused into the LiDAR stream by residual-based fusion modules. Finally, we introduce perception-aware losses into the optimization of the network.

\subsection{Pipeline of data pre-processing}
\label{sec:projection}
Existing methods~\cite{Madawy2019RGBAL,vora2020pointpainting} mainly project images to the LiDAR coordinate system using spherical projection. However, due to the sparse nature of point clouds, most of the appearance information from the images is lost with spherical projection (see Figure~\ref{fig:iv_vs_sv}). To address this issue, we propose perspective projection to project the sparse point clouds to the camera coordinate system.

\noindent\textbf{Formulation of perspective projection.}
Let $\{\bP, \bX, \by \}$ be one of the training samples from a given data set, where $\bP \in\mmR^{4\times N}$ indicates a point cloud from LiDAR and $N$ denotes the number of points. Each point $\bP_i$ in point cloud $\bP$ consists of 3D coordinates $(x, y, z)$ and a reflectance value $(r)$. Let $\bX \in \mmR^{3\times H\times W}$ be an image from an RGB camera, where $H$ and $W$ represent the height and width of the image, respectively. $\by\in \mmR^{N}$ is the set of semantic labels for point cloud $\bP$.

In perspective projection, we aim to project the point cloud $\bP$ from LiDAR coordinate to the camera coordinate to obtain the 2D LiDAR features $\widetilde{\bX}\in \mmR^{C\times H \times W}$. Here, $C$ indicates the number of channels~\wrt~the projected point cloud. Following~\cite{geiger2013vision}, we obtain $\bP_i=(x, y, z, 1)^\top$ by appending a fourth column to $\bP_i$ and compute the projected point $\widetilde{\bP}_i=(\widetilde{x},\widetilde{y}, \widetilde{z})^\top$ in the camera coordinates by
\begin{equation}
\label{eq:mapping_trans}
    \widetilde{\bP}_i = \bT \bR \bP_i,
\end{equation}
where $\bT\in \mmR^{3\times 4} $ is the projection matrix from LiDAR coordinates to camera coordinates. $\bR \in \mmR^{4\times 4}$ is expanded from the rectifying rotation matrix $\bR^{(0)}\in \mmR^{3\times 3}$ by appending a fourth zero row and column and setting $\bR(4, 4)=1$. The calibration parameters $\bT$ and $\bR^{(0)}$ can be obtained by the approach in~\cite{geiger2012automatic}. Subsequently, the corresponding pixel $(h, w)$ in the projected image~$\widetilde{\bX}$~\wrt~the point $\bP_i$ is computed by $h=\widetilde{x} / \widetilde{z}$ and $w=\widetilde{y} / \widetilde{z}$.

\begin{algorithm}[t]
    \caption{General Scheme of EPMF}
    \begin{algorithmic}[1]
        \REQUIRE Training data $\{\bP,\bX,\by\}$, TSNet with submodels $M,\widetilde{M}$, hyperparameters $\tau,\lambda,\gamma$.
        \WHILE{\textit{not convergent}}
            \STATE Project the point clouds $\bP$ by using perspective projection to obtain $\widetilde{\bX}$.
            \STATE Use $\{\widetilde{\bX},\bX\}$ as the inputs of TSNet and compute the output probabilities $\{\widetilde{\bO},\bO\}$ with Eq. (\ref{eq:output_tsnet}).
            \STATE Compute the perceptual confidence $\widetilde{\bC}$ and $\bC$.
            \STATE Construct perception-aware losses to measure the perceptual difference with Eqs. (\ref{eq:lidar_percept}) and (\ref{eq:camera_percept}).
            \STATE Update $\widetilde{M}$ and $M$ by minimizing the objective in Eq. (\ref{eq:tsnet_obj}).
        \ENDWHILE
    \end{algorithmic}
    \label{alg:optimization}
\end{algorithm}
\noindent\textbf{Cross-modal alignment and cropping.}
As shown in Figure~\ref{fig:arch_overview}(a), since we only focus on the segmentation of point clouds, directly projecting point clouds to the view of cameras leads to unnecessary computational costs. To address this issue, we introduce cross-modal alignment and cropping (CAC). First, we align the RGB image and the projected point clouds to find the overlap of the multi-modal inputs. Then, we crop both RGB images and projected point clouds to obtain compact inputs: For RGB images, we only keep the area that contains point clouds. For the projected point clouds, as the area outside the horizontal field of view (FOV) of the camera is covered by other cameras, we only keep the points within the horizontal FOV of the camera. In the case that the LiDAR sensor has a larger vertical FOV, we can keep the point clouds outside the image.\footnote{More discussions of CAC can be found in Seciton~\ref{sec:eff_improved}}

After applying CAC, we compute the features of the projected point clouds. Because the point cloud is very sparse, each pixel in the projected $\widetilde{\bX}$ may not have a corresponding point $\bp$. Therefore, we first initialize all pixels in $\widetilde{\bX}$ to 0. Following~\cite{cortinhal2020salsanext,milioto2019rangenet++}, we compute 5-channel LiDAR features,~\ie, $(d, x, y, z, r)$, for each pixel $(h, w)$ in the projected 2D image $\widetilde{\bX}$, where $d=\sqrt{x^2+y^2+z^2}$ represents the range value of each point. Thus, we set the number of channels $C$ to 5 in this work.

\subsection{Architecture design of EPMF}
\label{sec:arch}

As images and point clouds are different-modality data, it is difficult to handle both types of information from the two modalities by using a single network~\cite{krispel2020fuseseg}. 
Motivated by~\cite{feichtenhofer2016convolutional,simonyan2014two}, we propose a two-stream network (TSNet) that contains a camera stream and a LiDAR stream to process the features from camera and LiDAR, separately, as illustrated in Figure~\ref{fig:arch_overview}. In this way, we can use the network architectures designed for images and point clouds as the backbones of each stream in TSNet. 

\begin{figure}[t]
    \centering
    \includegraphics[width=\linewidth]{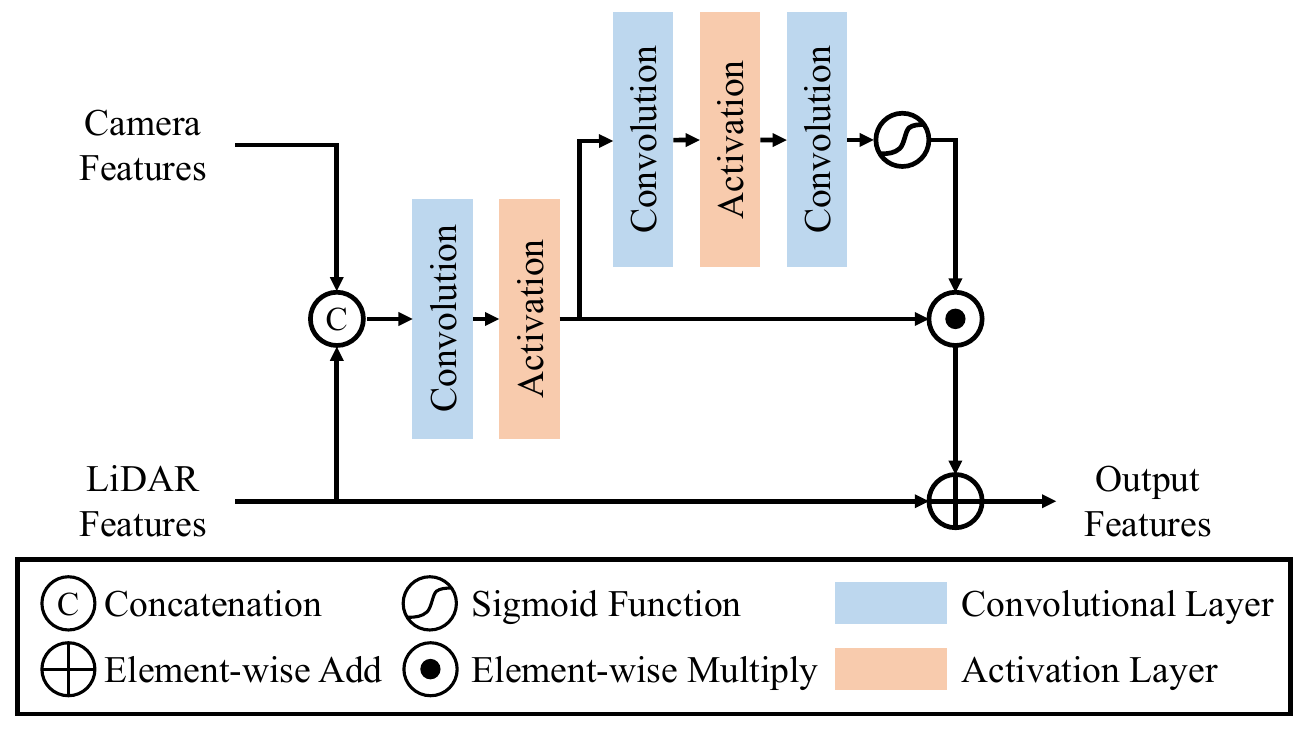}
    \caption{Illustration of the residual-based fusion (RF) module. RF fuses features from both the camera and LiDAR to generate the complementary information of the original LiDAR features.}
    \label{fig:fusion_module}
\end{figure}

\noindent\textbf{Formulation of Two Stream Network.}
Let $\widetilde{M}$ and $M$ be the LiDAR stream and the camera stream in TSNet, respectively. Let $\widetilde{\bO}\in \mmR^{S\times H \times W}$ and $\bO \in \mmR^{S\times H \times W}$ be the output probabilities~\wrt~each network, where $S$ indicates the number of semantic classes. The outputs of TSNet are computed by
\begin{equation}
\label{eq:output_tsnet}
    \left\{
    \begin{aligned}
        & \bO  = M(\bX), \\
        & \widetilde{\bO} = \widetilde{M}(\widetilde{\bX}).
    \end{aligned}
    \right.
\end{equation}

\noindent\textbf{Design of Residual-based Fusion Module.}
Since the features of images contain many details of objects, we then introduce a residual-based fusion module, as illustrated in Figure~\ref{fig:fusion_module}, to fuse the image features to the LiDAR stream\footnote{We discuss different designs of fusion modules in Section 2 of the supplementary material of~\cite{zhuang2021perception}}. Let $\{\bF_l\in \mmR^{C_l\times H_l \times W_l}\}_{l=1}^L$ be a set of image features from the camera stream, where $l$ indicates the layer in which we obtain the features. $C_l$ indicates the number of channels of the $l$-th layer in the camera stream. $H_l$ and $W_l$ indicate the height and width of the feature maps from the $l$-th layer, respectively. Let $\{\widetilde{\bF}_l\in \mmR^{\widetilde{C}_l\times H_l \times W_l}\}_{l=1}^L$ be the features from the LiDAR stream, where $\widetilde{C}_l$ indicates the number of channels of the $l$-th layer in the LiDAR stream. To obtain the fused features, we first concatenate the features from each network and use a convolutional layer to reduce the number of channels of the fused features. The fused features $\bF_l^{fuse}\in \mmR^{\widetilde{C}_l\times H_l \times W_l}$ are computed by 
\begin{equation}
\label{eq:fuse_cat}
    \bF_l^{fuse}= f_l([\widetilde{\bF}_l; \bF_l]),
\end{equation} 
where $[\cdot;\cdot]$ indicates the concatenation operation. $f_l(\cdot)$ is the convolution operation~\wrt~the $l$-th fusion module. 

Considering that the camera is easily affected by different lighting and weather conditions, the information from RGB images is not reliable in an outdoor environment. We use the fused features as the complement of the original LiDAR features and design the fusion module based on the residual structure~\cite{he2016deep}. Incorporating with the attention module~\cite{bochkovskiy2020yolov4}, the output features $\bF_l^{out}\in\mmR^{\widetilde{C}_l \times H_l \times W_l}$ of the fusion module are computed by
\begin{equation}
\label{eq:fuse_add}
    \bF_l^{out} = \widetilde{\bF}_l + \sigma(g_l(\bF_l^{fuse}))\odot \bF_l^{fuse},
\end{equation}
where $\sigma(x)=1/(1+e^{-x})$ indicates sigmoid function. $g_l(\cdot)$ indicates convolution operation in the attention module~\wrt~the $l$-th fusion module. $\odot$ indicates element-wise multiplication operation.

\subsection{Construction of perception-aware loss}
\label{sec:loss}

The construction of perception-aware loss is very important in our method. As demonstrated in Figure~\ref{fig:perceptual_confidence}, because the point clouds are very sparse, the LiDAR stream network learns only the local features of points while ignoring the shape of objects. In contrast, the camera stream can easily capture the shape and texture of objects from dense images. In other words, the perceptual features captured by the camera stream and LiDAR stream are different. With this intuition, we introduce a perception-aware loss to make the fusion network focus on the perceptual features from the camera and LiDAR.

To measure the perceptual confidence of the predictions~\wrt~the LiDAR stream, we first compute the entropy map $\widetilde{\bE} \in \mmR^{H \times W}$ by 
\begin{equation}
\label{eq:entropy}
    \widetilde{\bE}_{h,w} = - \frac{1}{\log{S}}\sum_{s=1}^{S}\widetilde{\bO}_{s,h,w} \log(\widetilde{\bO}_{s,h,w}).
\end{equation}
Following~\cite{renyi1961measures}, we use $\log{S}$ to normalize the entropy to $(0,1]$. Then, the perceptual confidence map $\widetilde{\bC}$~\wrt~the LiDAR stream is computed by $\widetilde{\bC} = \textbf{1}-\widetilde{\bE}$. For the camera stream, the confidence map is computed by $\bC=\textbf{1}-\bE$.

Note that not all information from the camera stream is useful. For example, the camera stream is confident inside objects but may make mistakes at the boundary of the objects. In addition, the predictions with lower confidence scores are more likely to be wrong. Incorporating a confidence threshold, we measure the importance of perceptual information from the camera stream by
\begin{equation}
\label{eq:lidar_weight}
    \widetilde{\mathbf{\Omega}}_{h,w} = \left\{
    \begin{array}{ll}
       \text{max}(\bC_{h,w} - \widetilde{\bC}_{h,w}, 0), & \text{if}~~\bC_{h,w} >\tau, \\
        0, &  \text{otherwise}. \\
    \end{array}
    \right.
\end{equation}
Here $\tau$ indicates the confidence threshold.

Inspired by~\cite{hinton2015distilling,jaritz2020xmuda,zhang2018deep}, to learn the perceptual information from the camera stream, we construct the perception-aware loss~\wrt~the LiDAR stream by 
\begin{equation}
\label{eq:lidar_percept}
    \widetilde{\mL}_{per} = \frac{1}{Q}\sum_{h=1}^{H}\sum_{w=1}^{W}\widetilde{\mathbf{\Omega}}_{h,w} D_{KL}(\widetilde{\bO}_{:,h,w}|| \bO_{:,h,w}),
\end{equation}
where $Q=H \cdot W$ and $D_{KL}(\cdot||\cdot)$ indicates the Kullback-Leibler divergence~\cite{hinton2015distilling}.

For the camera stream, the importance of information from LiDAR stream is computed by
\begin{equation}
\label{eq:camera_weight}
    \mathbf{\Omega}_{h,w} = \left\{
    \begin{array}{ll}
        \text{max}( \widetilde{\bC}_{h,w} - \bC_{h,w},0), &  \text{if}~~\widetilde{\bC}_{h,w} >\tau, \\
        0, &  \text{otherwise}. \\
    \end{array}
    \right.
\end{equation}
The perception-aware loss~\wrt~the camera stream is
\begin{equation}
\label{eq:camera_percept}
    \mL_{per} = \frac{1}{Q}\sum_{h=1}^{H}\sum_{w=1}^{W}\mathbf{\Omega}_{h,w} D_{KL}(\bO_{:,h,w}||\widetilde{\bO}_{:,h,w}).
\end{equation}

\subsection{Objective functions}
In addition to the perception-aware loss, we also use multi-class focal loss~\cite{lin2017focal} and Lov{\'a}sz-softmax loss~\cite{berman2018lovasz}, which are commonly used in existing segmentation work~\cite{cortinhal2020salsanext,zhu2021cylindrical}, to train the two stream network.

Let $\bY\in\mmR^{H\times W}$ be the projected labels in the camera coordinates. $H$ and $W$ indicate the height and width, respectively. For each point $\bP_i$, we project the 3D coordinates $(x,y,z)$ to the pixel $(h,w)$ in the camera coordinate system by using perspective projection. Then, we initialize all pixels in $\bY$ by 0 and compute the projected labels in $\bY$ by 
\begin{equation}
    \bY_{h,w} := \by_i.
\end{equation}

 The multi-class focal loss~\wrt~the LiDAR stream is defined as
\begin{equation}
    \widetilde{\mL}_{foc} =\frac{1}{K}\sum_{s=1}^{S}\sum_{h=1}^{H}\sum_{w=1}^{W}\alpha_s\mathbbm{1}\{\bY_{h,w}=s\} FL(\widetilde{\bO}_{s,h,w}),
\end{equation}
where $FL(p)=-(1-p)^2\log(p)$ denotes the focal-loss function. $\alpha_s$ denotes the weights~\wrt the $s$-th class. $K=\sum_{s=1}^{S}\sum_{h=1}^{H}\sum_{w=1}^{W}\mathbbm{1}\{\bY_{h,w}=s\}$ indicates the number of available labels. $\mathbbm{1}\{\cdot\}$ indicates the indicator function. Then, the multi-class focal loss~\wrt~the camera stream is
\begin{equation}
    \mL_{foc} =\frac{1}{K}\sum_{s=1}^{S}\sum_{h=1}^{H}\sum_{w=1}^{W}\mathbbm{1}\{\bY_{h,w}=s\} FL(\bO_{s,h,w}),
\end{equation}

\begin{figure}[t]
    \centering
    \includegraphics[width=1.0\linewidth]{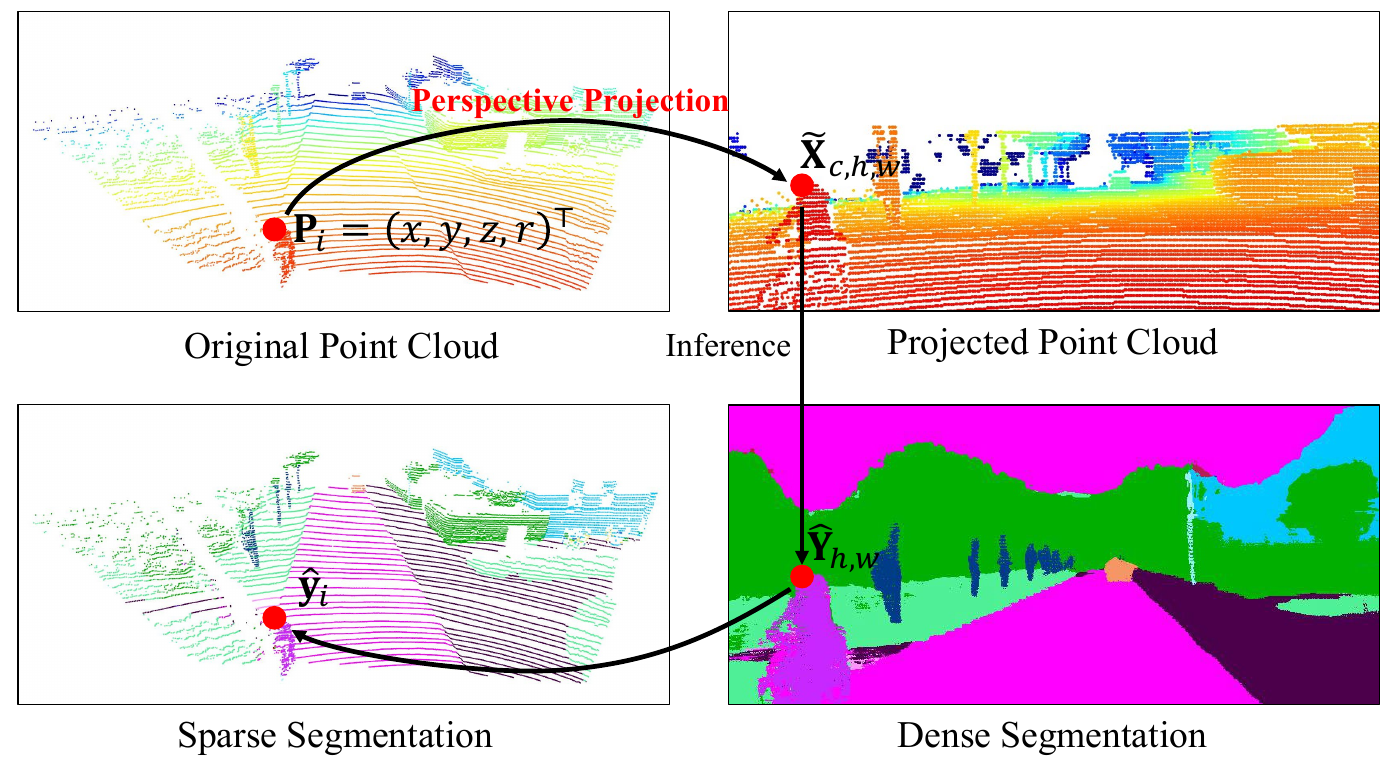}
    \caption{Illustration of the pipeline to obtain the sparse segmentation from the dense prediction results. $\widetilde{\bX}$ indicates the projected point cloud. $\widehat{\bY}$ and $\widehat{\by}$ indicate the dense predictions and sparse predictions, respectively. For each point $\bP_i$, we first compute the corresponding pixel $(h,w)$ in the camera coordinate system by perspective projection. Second, we get the dense segmentation $\widehat{\bY}$ from the prediction results of PMF. Last, we obtain the corresponding sparse prediction $\widehat{\by}_i$~\wrt~the point $\bP_i$ from the dense segmentation $\widehat{\bY}_{h,w}$.}
    \label{fig:rela_sparse_dense}
\end{figure}

The Lov{\'a}sz-softmax loss~\wrt~the LiDAR stream is
\begin{equation}
\label{eq:lov_loss}
    \widetilde{\mL}_{lov}=\frac{1}{S}\sum_{s=1}^{S}\overline{\Delta_{J_{s}}}(\widetilde{\bm}(s)),
\end{equation}
where
\begin{equation}
\widetilde{\bm}_{i}(s)=\left\{\begin{array}{ll}
1-\widetilde{\bO}_{s,h,w}  & \text {if}~~s=\bY_{h,w}, \\
\widetilde{\bO}_{s,h,w}  & \text {otherwise}.
\end{array}\right.
\end{equation}
$\overline{\Delta_{J_{s}}}$ indicates the Lov{\'a}sz extension of the Jaccard index for class $s$. Here, $(h,w)$ is obtained from the 3D coordinates $(x,y,z)$ of $\bP_i$ by using perspective projection. $\widetilde{\bm}(s)\in [0, 1]^N$ indicates the vector of errors. The Lov{\'a}sz-softmax loss~\wrt~the camera stream is defined as 
\begin{equation}
    \mL_{lov}=\frac{1}{S}\sum_{s=1}^{S}\overline{\Delta_{J_{s}}}(\bm(s)),
\end{equation}
where 
\begin{equation}
\bm_i(s)=\left\{\begin{array}{ll}
1-\bO_{s,h,w}  & \text {if}~~s=\bY_{h,w}, \\
\bO_{s,h,w}  & \text {otherwise}.
\end{array}\right.
\end{equation}

By considering the objective functions of both LiDAR stream and camera stream, we formulate the objective function of the proposed two-stream network as

\begin{equation}
\label{eq:tsnet_obj}
    \mL = \widetilde{\mL}_{foc} + \mL_{foc}  + \widetilde{\lambda}\widetilde{\mL}_{lov} + \lambda\mL_{lov} + \widetilde{\gamma}\widetilde{\mL}_{per} + \gamma\mL_{per},
\end{equation}
where $\lambda,\widetilde{\lambda},\gamma$ and $\widetilde{\gamma}$ indicate the hyper-parameters that balance different losses.

\subsection{Pipeline of post-processing}
With the proposed perception-aware losses, PMF generates dense segmentation results with information from RGB images and point clouds. We then obtain the sparse prediction from the dense results. Let $\widetilde{\bO}\in\mmR^{S\times H\times W}$ be the output probabilities of the LiDAR stream. $S$ indicates the number of classes. $H$ and $W$ indicate the height and width of the predictions, respectively. Let $\widehat{\bY}\in \mmR^{H\times W}$ be the dense predictions from the LiDAR stream. Then, the dense predictions are computed by 
\begin{equation}
    \widehat{\bY}_{h,w} = \arg\max_s \widetilde{\bO}_{s,h,w}.
\end{equation}

Let $\widehat{\by}\in \mmR^{N}$ be the sparse predictions of point cloud $\bP$. As shown in Figure~\ref{fig:rela_sparse_dense}, for each point $\bP_i$, we first project the 3D coordinates $(x,y,z)$ to the camera coordinate system by using perspective projection and compute the corresponding pixel $(h,w)$ in the projected image. Then the semantic prediction $\widehat{\by}_i$~\wrt~the point $\bP_i$ is computed by
\begin{equation}
    \widehat{\by}_i := \widehat{\bY}_{h,w}.
\end{equation}

Note that for the point cloud with multi-camera views,~\eg, nuScenes, there are overlaps between different camera views. To address this issue, we conduct inference for each camera view and merge the results by assigning the predictions with the highest confidence scores to the points in the overlaps of different views.

\subsection{Techniques to improve efficiency and effectiveness}
On top of the two-stream network designed in~\cite{zhuang2021perception}, we further explore the techniques to improve the efficiency and effectiveness of the fusion network\footnote{We study the effect of the proposed techniques in Section~\ref{sec:eff_improved}}. 

\noindent\textbf{Dropping decoder of camera stream.} By introducing the perception-aware loss, the knowledge of the camera stream is distilled into the LiDAR stream. Therefore, we only use the predictions of LiDAR stream while dropping the results of the camera stream during inference. In this sense, we can also drop the decoder of the camera stream to speed up inference. 

\noindent\textbf{Improved contextual module.} In~\cite{cortinhal2020salsanext}, Cortinhal~\etal have designed the contextual module that improves the ability of SalsaNet~\cite{aksoy2019salsanet} for comprehending global contextual information. However, the proposed contextual module is explored under spherical projection and may be ineffective with perspective projection. In our experiments, we find that the high resolution of the projected point cloud feature is unnecessary due to the sparse nature of point clouds. Therefore, we insert extra down-sampling operation into the contextual module of LiDAR stream to reduce the resolution of point cloud features and improve the efficiency of LiDAR stream. Moreover, to reduce the impact of the sparse issue of projected point clouds, we replace the convolutional layers in the contextual modules of LiDAR stream with sparse invariant convolutional layers~\cite{uhrig2017sparsity}. 

\noindent\textbf{Fusing high-level LiDAR features.} In the two-stream network, the camera stream generates predictions with RGB images only, which, however, results in the unsatisfactory segmentation performance of the camera stream and may limit in performance of the fusion network. To address this issue, we further fuse the high-level features from the last stage of the LiDAR stream backbone into the camera stream by a concatenate operation. In this way, we improve the performance of the camera stream without introducing extra computational costs and thus boost the effectiveness of the perception-aware loss. 

\begin{table}
\centering
\caption{Comparisons of the number of points of SemanticKITTI and SemanticKITTI-FV.}
  \scalebox{1.0}{
    \begin{tabular}{c|cc|c}
    \hline
    Dataset splits & SemanticKITTI & SemanticKTTI-FV & Percentage \\
    \hline
    
    Training set & $2.35\times 10^9$ & $3.78\times 10^8$& 16.03\% \\ 
    Validation set & $4.99\times 10^8$& $8.00\times 10^7$& 16.07\%\\ 
    \hline
    \end{tabular}
    }

\label{tab:dataset_smk}
\end{table}

\begin{table}
\centering
\caption{Comparisons of the number of valid labels of SemanticKITTI and SemanticKITTI-FV.}
  \scalebox{1.0}{
    \begin{tabular}{c|cc|c}
    \hline
    Dataset splits & SemanticKITTI & SemanticKTTI-FV & Percentage \\
    \hline
    
    Training set & $2.28\times 10^9$ & $3.63\times 10^8$& 15.93\% \\ 
    Validation set & $4.77\times 10^8$& $7.57\times 10^7$& 15.88\%\\ 
    \hline
    \end{tabular}
    }

\label{tab:dataset_smk_labels}
\end{table}

\begin{table*}[t]

\centering
\caption{Comparisons on SemanticKITTI-FV validation set. \textbf{L} indicates LiDAR-only methods. \textbf{L+C} indicates fusion-based methods. * indicates the results based on our implementation. The \textbf{bold} numbers indicate the best results.}

  \scalebox{0.82}{
  \begin{threeparttable}
\begin{tabular}{l|c|ccccccccccccccccccc|c}
\hline
Method & \rotatebox{90}{Modality}  & \rotatebox{90}{Car} & \rotatebox{90}{Bicycle} & \rotatebox{90}{Motorcycle} & \rotatebox{90}{Truck} & \rotatebox{90}{Other-vehicle} & \rotatebox{90}{Person} & \rotatebox{90}{Bicyclist} & \rotatebox{90}{Motorcyclist} & \rotatebox{90}{Road} & \rotatebox{90}{Parking} & \rotatebox{90}{Sidewalk} & \rotatebox{90}{Other-ground} & \rotatebox{90}{Building} & \rotatebox{90}{Fence} & \rotatebox{90}{Vegetation} & \rotatebox{90}{Trunk} & \rotatebox{90}{Terrain} & \rotatebox{90}{Pole} & \rotatebox{90}{Traffic-sign} & \rotatebox{90}{mIoU (\%)} \\ 

\hline\hline
RandLANet~\cite{hu2020randla} & L 
& 92.0 & 8.0 & 12.8 & 74.8 & 46.7 & 52.3 & 46.0 & 0.0 & 93.4 & 32.7 & 73.4 & 0.1 & 84.0 & 43.5 & 83.7 & 57.3 & 73.1 & 48.0 & 27.3 & 50.0 \\ 
RangeNet++~\cite{milioto2019rangenet++} & L 
& 89.4 & 26.5 & 48.4 & 33.9 & 26.7 & 54.8 & 69.4 & 0.0 & 92.9 & 37.0 & 69.9 & 0.0 & 83.4 & 51.0 & 83.3 & 54.0 & 68.1 & 49.8 & 34.0 & 51.2 \\
SequeezeSegV2~\cite{wu2019squeezesegv2} & L 
& 82.7 & 15.1 & 22.7 & 25.6 & 26.9 & 22.9 & 44.5 & 0.0 & 92.7 & 39.7 & 70.7 & 0.1 & 71.6 & 37.0 & 74.6 & 35.8 & 68.1 & 21.8 & 22.2 & 40.8 \\ 
SequeezeSegV3~\cite{xu2020squeezesegv3} & L 
& 87.1 & 34.3 & 48.6 & 47.5 & 47.1 & 58.1 & 53.8 & 0.0 & 95.3 & 43.1 & 78.2 & 0.3 & 78.9 & 53.2 & 82.3 & 55.5 & 70.4 & 46.3 & 33.2 & 53.3 \\ 

SalsaNext~\cite{cortinhal2020salsanext} & L 
& 90.5 & 44.6 & 49.6 & 86.3 & 54.6 & 74.0 & 81.4 & 0.0 & 93.4 & 40.6 & 69.1 & 0.0 & 84.6 & 53.0 & 83.6 & 64.3 & 64.2 & 54.4 & 39.8 & 59.4 \\ 

MinkowskiNet~\cite{choy20194d} & L 
& 95.0 & 23.9 & 50.4 & 55.3 & 45.9 & 65.6 & 82.2 & 0.0 & 94.3 & 43.7 & 76.4 & 0.0 & 87.9 & 57.6 & 87.4 & 67.7 & 71.5 & 63.5 & 43.6 & 58.5 \\ 

SPVNAS~\cite{tang2020searching} & L 
& 96.5 & 44.8 & 63.1 & 59.9 & 64.3 & 72.0 & 86.0 & 0.0 & 93.9 & 42.4 & 75.9 & 0.0 & 88.8 & 59.1 & 88.0 & 67.5 & 73.0 & 63.5 & 44.3 & 62.3 \\ 

Cylinder3D~\cite{zhu2021cylindrical} & L 
& 96.4 & 61.5 & 78.2 & 66.3 & 69.8 & 80.8 & 93.3 & 0.0 & 94.9 & 41.5 & 78.0 & 1.4 & 87.5 & 50.0 & 86.7 & 72.2 & 68.8 & 63.0 & 42.1 & 64.9 \\ 

2DPASS~\cite{yan20222dpass} & L+C 
& 96.6 & 60.8 & 71.6 & 82.0 & 77.8 & 78.2 & 92.0 & 0.2 & 93.9 & 44.4 & 75.7 & 3.1 & 89.5 & 60.9 & 88.7 & 72.6 & 74.0 & 61.5 & 45.5 & \textbf{66.8}\\ 

2DPASS\tnote{1}~\cite{yan20222dpass} & L+C 
& 93.6 & 54.9 & 71.6 & 81.6 & 43.5 & 71.5 & 82.2 & 0.2 & 94.0 & 33.8 & 75.1 & 0.2 & 88.7 & 57.3 & 88.5 & 67.1 & 75.1 & 56.0 & 40.2 & 61.8 \\

PointPainting*~\cite{vora2020pointpainting} & L+C 
& 94.7 & 17.7 & 35.0 & 28.8 & 55.0 & 59.4 & 63.6 & 0.0 & 95.3 & 39.9 & 77.6 & 0.4 & 87.5 & 55.1 & 87.7 & 67.0 & 72.9 & 61.8 & 36.5 & 54.5\\

RGBAL*~\cite{Madawy2019RGBAL} & L+C 
& 87.3 & 36.1 & 26.4 & 64.6 & 54.6 & 58.1 & 72.7 & 0.0 & 95.1 & 45.6 & 77.5 & 0.8 & 78.9 & 53.4 & 84.3 & 61.7 & 72.9 & 56.1 & 41.5 & 56.2 \\
PMF\cite{zhuang2021perception} & L+C 
& 95.4 & 47.8 & 62.9 & 68.4 & 75.2 & 78.9 & 71.6 & 0.0 & 96.4 & 43.5 & 80.5 & 0.1 & 88.7 & 60.1 & 88.6 & 72.7 & 75.3 & 65.5 & 43.0 & 63.9
\\
\hline
Uni-modal baseline & L & 93.2 & 35.9 & 33.9 & 78.2 & 54.4 & 63.1 & 71.9 & 0.0 & 95.0 & 42.4 & 80.0 & 0.7 & 86.0 & 53.7 & 86.1 & 63.5 & 73.7 & 59.4 & 41.9 & 58.6 \\
EPMF (Ours) & L+C
& 95.4 & 52.5 & 66.3 & 82.4 & 80.5 & 77.2 & 85.1 & 0.0 & 95.8 & 48.1 & 80.5 & 0.8 & 89.2 & 66.1 & 86.0 & 70.0 & 68.9 & 62.6 & 43.6 & 65.9 \\
\hline
\end{tabular}
\begin{tablenotes}
    \item[1] Model trained under our settings. 
\end{tablenotes}
\end{threeparttable}
}

\label{tab:semanti_kitti_results}
\end{table*}
     
\section{Experiments}

In this section, we first compare EPMF with the state-of-the-art methods on the benchmark data sets. Then, we provide distance-based evaluation and qualitative results of our methods. Last, we conduct experiments to evaluate the efficiency of our method.

We organize the experiments as follows. 
1) We introduce the benchmark data sets and evaluation metrics in Section~\ref{sec:exp_dataset}. 
2) We provide the implementation details of our method in Section~\ref{sec:implementation}. 
3) We evaluate the performance of our method on several benchmark data sets in Section~\ref{sec:main_results}. 
4) We investigate the performance of our method under different distances in Section~\ref{sec:distance_eval}. 
5) We discuss the efficiency of the proposed method in Section~\ref{sec:efficiency}. 
6) We show the qualitative results in Section~\ref{sec:qualitative_eval}. 
7) We study the robustness of the proposed method on adversarial samples in Section~\ref{sec:abla_anti_attack}.

\subsection{Data sets and Evaluation Metrics}
\label{sec:exp_dataset}
\subsubsection{Data sets}
We empirically evaluate our method on several benchmark data sets, including SemanticKITTI-FV~\cite{behley2019semantickitti}, nuScenes~\cite{caesar2020nuscenes}, and A2D2~\cite{geyer2020a2d2}. 

SemanticKITTI is a large-scale data set based on the KITTI Odometry Benchmark~\cite{geiger2012we}, providing 43,000 scans with point-wise semantic annotation, where 21,000 scans (sequence 00-10) are available for training and validation. The data set has 19 semantic classes for the evaluation of semantic benchmarks. The point clouds are collected using a Velodyne HDL-64E sensor, which has 64 beams vertically. Since SemanticKITTI provides only the images of the front-view camera, we project the point clouds to a perspective view and keep only the available points on the images to build a subset of SemanticKITTI, namely, SemanticKITTI-FV. The comparisons between SemanticKITTI and SemanticKITTI-FV are shown in Table~\ref{tab:dataset_smk} and Table~\ref{tab:dataset_smk_labels}. SemanticKITTI-FV has only 15.93\% data for training compared with the full SemanticKITTI data set.

nuScenes contains 1,000 driving scenes with different weather and light conditions. The scenes are split into 28,130 training frames and 6,019 validation frames, which are collected with a Velodyne HDL-32E sensor. Unlike SemanticKITTI, which provides only the images of the front-view camera, nuScenes has 6 cameras for different views of LiDAR.

A2D2 provides 38-class semantic segmentation images and point cloud labels for 41,277 non-sequential frames, which are collected with six cameras and five Velodyne VLP-16 sensors. Following~\cite{zhang2020polarnet}, we split 22,408 scans for training, 2,274 for validation, and 13,264 for testing, respectively. Similar to nuScenes, the sensor suite of A2D2 provides full 360° coverage. However, the camera and LiDAR sensor orientations are optimized manually to minimize the blind spot around the vehicle and maximize camera and LiDAR field of view overlap, which makes A2D2 lack horizontal scan lines.

\begin{table*}[t]
\centering
\caption{Comparisons on the nuScenes validation set. \textbf{L} indicates LiDAR-only methods. \textbf{L+C} indicates fusion-based methods. $^\dagger$ indicates the results with test-time augmentation. The \textbf{bold} numbers indicate the best results.}

  \scalebox{0.93}{
\begin{tabular}{l|c|cccccccccccccccc|c}
\hline
Method & \rotatebox{90}{Modality} & \rotatebox{90}{Barrier} & \rotatebox{90}{Bicycle} & \rotatebox{90}{Bus} & \rotatebox{90}{Car} & \rotatebox{90}{Construction} & \rotatebox{90}{Motorcycle} & \rotatebox{90}{Pedestrian} & \rotatebox{90}{Traffic-cone} & \rotatebox{90}{Trailer} & \rotatebox{90}{Truck} & \rotatebox{90}{Driveable} & \rotatebox{90}{Other-flat} & \rotatebox{90}{Sidewalk} & \rotatebox{90}{Terrain} & \rotatebox{90}{Manmade} & \rotatebox{90}{Vegetation} & \rotatebox{90}{mIoU (\%)} \\ 
\hline\hline
RangeNet++~\cite{milioto2019rangenet++} & L
& 66.0 & 21.3 & 77.2 & 80.9 & 30.2 & 66.8 & 69.6 & 52.1 & 54.2 & 72.3 & 94.1 & 66.6 & 63.5 & 70.1 & 83.1 & 79.8 & 65.5 \\ 
PolarNet~\cite{zhang2020polarnet} & L
& 74.7 & 28.2 & 85.3 & 90.9 & 35.1 & 77.5 & 71.3 & 58.8 & 57.4 & 76.1 & 96.5 & 71.1 & 74.7 & 74.0 & 87.3 & 85.7 & 71.0 \\ 
Salsanext~\cite{cortinhal2020salsanext} & L 
& 74.8 & 34.1 & 85.9 & 88.4 & 42.2 & 72.4 & 72.2 & 63.1 & 61.3 & 76.5 & 96.0 & 70.8 & 71.2 & 71.5 & 86.7 & 84.4 & 72.2 \\ 

SVASeg~\cite{zhao2022svaseg} & L
& 73.1 & 44.5 & 88.4 & 86.6 & 48.2 & 80.5 & 77.7 & 65.6 & 57.5 & 82.1 & 96.5 & 70.5 & 74.7 & 74.6 & 87.3 & 86.9 & 74.7\\

PVKD~\cite{hou2022point} & L
& 76.2 & 40.0 & 90.2 & 94.0 & 50.9 & 77.4 & 78.8 & 64.7 & 62.0 & 84.1 & 96.6 & 71.4 & 76.4 & 76.3 & 90.3 & 86.9 & 76.0 \\
AMVNet~\cite{liong2020amvnet} & L
& 79.8 & 32.4 & 82.2 & 86.4 & 62.5 & 81.9 & 75.3 & 72.3 & 83.5 & 
65.1 & 97.4 & 67.0 & 78.8 & 74.6 & 90.8 & 87.9 & 76.1 \\
Cylinder3D~\cite{zhu2021cylindrical} & L 
& 76.4 & 40.3 & 91.3 & 93.8 & 51.3 & 78.0 & 78.9 & 64.9 & 62.1 & 84.4 & 96.8 & 71.6 & 76.4 & 75.4 & 90.5 & 87.4 & 76.1 \\ 
RPVNet~\cite{xu2021rpvnet} & L
& 78.2 & 43.4 & 92.7 & 93.2 & 49.0 & 85.7 & 80.5 & 66.0 & 66.9 & 
84.0 & 96.9 & 73.5 & 75.9 & 76.0 & 90.6 & 88.9 & 77.6 \\
SDSeg3D~\cite{li2022self} & L
& 77.5 & 49.4 & 93.9 & 92.5 & 54.9 & 86.7 & 80.1 & 67.8 & 65.7 & 
86.0 & 96.4 & 74.0 & 74.9 & 74.5 & 86.0 & 82.8 & 77.7 \\

SDSeg3D$^\dagger$~\cite{li2022self} & L
& 78.2 & 52.8 & 94.5 & 93.1 & 54.5 & 88.1 & 82.2 & 69.4 & 67.3 & 86.6 & 96.4 & 74.5 & 75.2 & 75.3 & 87.1 & 84.1 & 78.7 \\

WaffleIron~\cite{puy2023using} & L
& 78.7 & 51.3 & 93.6 & 88.2 & 47.2 & 86.5 & 81.7 & 68.9 & 69.3 & 83.1 & 96.9 & 74.3 & 75.6 & 74.2 & 87.2 & 85.2 & 77.6 \\

WaffleIron$^\dagger$~\cite{puy2023using} & L 
& 79.8 & 53.8 & 94.3 & 87.6 & 49.6 & 89.1 & 83.8 & 70.6 & 72.7 & 84.9 & 97.1 & 75.8 & 76.5 & 75.9 & 87.8 & 86.3 & 79.1 \\

RangeFormer~\cite{kong2023rethinking} & L
& 78.0 & 45.2 & 94.0 & 92.9 & 58.7 & 83.9 & 77.9 & 69.1 & 63.7 & 85.6 & 96.7 & 74.5 & 75.1 & 75.3 & 89.1 & 87.5 & 78.1 \\

SphereFormer~\cite{lai2023spherical} & L
& 77.7 & 43.8 & 94.5 & 93.1 & 52.4 & 86.9 & 81.2 & 65.4 & 73.4 & 85.3 & 97.0 & 73.4 & 75.4 & 75.0 & 91.0 & 89.2 & 78.4 \\

SphereFormer$^\dagger$~\cite{lai2023spherical} & L
& 78.7 & 46.7 & 95.2 & 93.7 & 54.0 & 88.9 & 81.1 & 68.0 & 74.2 & 86.2 & 97.2 & 74.3 & 76.3 & 75.8 & 91.4 & 89.7 & 79.5\\
2DPASS~\cite{yan20222dpass} & L+C
& 74.4 & 44.3 & 93.6 & 92.0 & 54.0 & 79.7 & 78.9 & 57.2 & 72.5 & 85.7 & 96.2 & 72.7 & 74.1 & 74.5 & 87.5 & 85.4 & 76.4 \\

2DPASS$^\dagger$~\cite{yan20222dpass} & L+C 
& 77.0 & 50.4 & 95.9 & 94.2 & 56.0 & 86.0 & 81.9 & 64.4 & 76.9 & 88.6 & 96.8 & 75.4 & 76.7 & 76.4 & 88.9 & 86.6 & 79.5\\
2D3DNet~\cite{genova2021learning} & L+C
& 78.3 & 55.1 & 95.4 & 87.7 & 59.4 & 79.3 & 80.7 & 70.2 & 68.2 & 86.6 & 96.1 & 74.9 & 75.7 & 75.1 & 91.4 & 89.9 & 79.0 \\
PMF~\cite{zhuang2021perception} & L+C
& 74.1 & 46.6 & 89.8 & 92.1 & 57.0 & 77.7 & 80.9 & 70.9 & 64.6 & 82.9 & 95.5 & 73.3 & 73.6 & 74.8 & 89.4 & 87.7 & 76.9 \\
PMF-R50~\cite{zhuang2021perception} & L+C 
& 74.9 & 55.4 & 91.0 & 93.0 & 60.5 & 80.3 & 83.2 & 73.6 & 67.2 & 84.5 & 95.9 & 75.1 & 74.6 & 75.5 & 90.3 & 89.0 & 79.0 \\
\hline
Uni-modal baseline & L & 
74.9 & 19.2 & 73.5 & 89.3 & 36.9 & 63.1 & 68.2 & 52.0 & 64.5 & 74.4 & 96.8 & 73.0 & 75.6 & 75.1 & 87.8 & 86.0 & 69.4 \\
EPMF (Ours) & L+C
& 79.7 & 55.8 & 96.0 & 92.4 & 65.6 & 86.4 & 80.9 & 74.3 & 68.1 & 87.0 & 97.0 & 75.6 & 76.2 & 76.3 & 90.2 & 88.1 & \textbf{80.6} \\

\hline
\end{tabular}
}

\label{tab:nuScenes_results}
\end{table*}

\begin{table*}[t]
\centering
\caption{Comparisons on the nuScenes test set. \textbf{L} indicates LiDAR-only methods. \textbf{L+C} indicates fusion-based methods. * represents the results reported in the leader board of nuScenes. $\dagger$ indicates the results with test-time augmentation. $\ddag$ denotes the results with additional fine-tuning with class re-sampling before the leaderboard submission. The \textbf{bold} numbers indicate the best results.}
\vspace{0.1in}
  \scalebox{0.93}{
\begin{tabular}{l|c|cccccccccccccccc|c}
\hline

Method & \rotatebox{90}{Modality} & \rotatebox{90}{Barrier} & \rotatebox{90}{Bicycle} & \rotatebox{90}{Bus} & \rotatebox{90}{Car} & \rotatebox{90}{Construction} & \rotatebox{90}{Motorcycle} & \rotatebox{90}{Pedestrian} & \rotatebox{90}{Traffic-cone} & \rotatebox{90}{Trailer} & \rotatebox{90}{Truck} & \rotatebox{90}{Driveable} & \rotatebox{90}{Other-flat} & \rotatebox{90}{Sidewalk} & \rotatebox{90}{Terrain} & \rotatebox{90}{Manmade} & \rotatebox{90}{Vegetation} & \rotatebox{90}{mIoU (\%)} \\ %
\hline\hline

PolarNet~\cite{zhang2020polarnet} & L
& 72.2 & 16.8 & 77.0 & 86.5 & 51.1 & 69.7 & 64.8 & 54.1 & 69.7 & 63.4 & 96.6 & 67.1 & 77.7 & 72.1 & 87.1 & 84.4 & 69.4 \\ 

AMVNet~\cite{liong2020amvnet} & L
& 79.8 & 32.4 & 82.2 & 86.4 & 62.5 & 81.9 & 75.3 & 72.3 & 83.5 & 65.1 & 97.4 & 67.0 & 78.8 & 74.6 & 90.8 & 87.9 & 76.1 \\
Cylinder3D*~\cite{zhu2021cylindrical} & L 
& 82.8 & 29.8 & 84.3 & 89.4 & 63.0 & 79.3 & 77.2 & 73.4 & 84.6 & 69.1 & 97.7 & 70.2 & 80.3 & 75.5 & 90.4 & 87.6 & 77.2 \\ 

SPVNAS*~\cite{tang2020searching} & L
& 80.0 & 30.0 & 91.9 & 90.8 & 64.7 & 79.0 & 75.6 & 70.9 & 81.0 & 74.6 & 97.4 & 69.2 & 80.0 & 76.1 & 89.3 & 87.1 & 77.4 \\
AF2S3Net*~\cite{cheng20212} & L
& 78.9 & 52.2 & 89.9 & 84.2 & 77.4 & 74.3 & 77.3 & 72.0 & 83.9 & 73.8 & 97.1 & 66.5 & 77.5 & 74.0 & 87.7 & 86.8 & 78.3 \\

RangeFormer~\cite{kong2023rethinking} & L 
& 83.9 & 46.1 & 89.4 & 89.2 & 70.3 & 83.3 & 75.4 & 72.5 & 81.4 & 71.1 & 95.6 & 68.5 & 77.3 & 73.4 & 89.3 & 86.9 & 78.3 \\
RangeFormer$^\dagger$~\cite{kong2023rethinking} & L 
& 85.6 & 47.4 & 91.2 & 90.9 & 70.7 & 84.7 & 77.1 & 74.1 & 83.2 & 72.6 & 97.5 & 70.7 & 79.2 & 75.4 & 91.3 & 88.9 & 80.1 \\

SphereFormer~\cite{lai2023spherical} & L
& 81.5 & 39.7 & 93.4 & 87.5 & 66.4 & 75.7 & 77.2 & 70.6 & 85.6 & 73.6 & 97.6 & 64.8 & 79.8 & 75.0 & 92.2 & 89.0 & 78.1 \\
SphereFormer$^{\dagger\ddag}$~\cite{lai2023spherical} & L
& 83.3 & 39.2 & 94.7 & 92.5 & 77.5 & 84.2 & 84.4 & 79.1 & 88.4 & 78.3 & 97.9 & 69.0 & 81.5 & 77.2 & 93.4 & 90.2 & \textbf{81.9} \\
2DPASS$^{\dagger\ddag}$~\cite{yan20222dpass} & L+C
& 81.7 & 55.3 & 92.0 & 91.8 & 73.3 & 86.5 & 78.5 & 72.5 & 84.7 & 75.5 & 97.6 & 69.1 & 79.9 & 75.5 & 90.2 & 88.0 & 80.8\\
2D3DNet$^\ddag$~\cite{genova2021learning} & L+C
& 83.0 & 59.4 & 88.0 & 85.1 & 63.7 & 84.4 & 82.0 & 76.0 & 84.8 & 71.9 & 96.9 & 67.4 & 79.8 & 76.0 & 92.1 & 89.2 & 80.0 \\   

PMF~\cite{zhuang2021perception} & L+C
& 80.1 & 35.7 & 79.7 & 86.0 & 62.4 & 76.3 & 76.9 & 73.6 & 78.5 & 66.9 & 97.1 & 65.3 & 77.6 & 74.4 & 89.5 & 87.7 & 75.5 \\
PMF-R50\cite{zhuang2021perception} & L+C
& 82.1 & 40.3 & 80.9 & 86.4 & 63.7 & 79.2 & 79.8 & 75.9 & 81.2 & 67.1 & 97.3 & 67.7 & 78.1 & 74.5 & 89.9 & 88.5 & 77.0 \\
\hline
EPMF (Ours) & L+C
& 76.9 & 39.8 & 90.3 & 87.8 & 72.0 & 86.4 & 79.6 & 76.6 & 84.1 & 74.9 & 97.7 & 66.4 & 79.5 & 76.4 & 91.1 & 87.9 & 79.2
\\

\hline
\end{tabular}
}
\label{tab:nus_test_set}
\end{table*}

\begin{table*}[t]
\centering
\caption{Comparisons on A2D2 test set - Part1. The \textbf{bold} numbers indicate the best results.}
  \scalebox{0.88}{
\begin{tabular}{l|ccccccccccccccccccc|c}
\hline
Method & \rotatebox{90}{Car} & \rotatebox{90}{Bicycle} & \rotatebox{90}{Pedestrian} & \rotatebox{90}{Truck} & \rotatebox{90}{Small-vehi} & \rotatebox{90}{Traffic-signal} & \rotatebox{90}{Traffic-signl} & \rotatebox{90}{Utility-vehi} & \rotatebox{90}{Sidebars} & \rotatebox{90}{Bumper} & \rotatebox{90}{Curbstone} & \rotatebox{90}{Solid line} & \rotatebox{90}{Irrelevant signs} & \rotatebox{90}{Road blocks} & \rotatebox{90}{Tractor} & \rotatebox{90}{Non-drivable} & \rotatebox{90}{Zebra crossing} & \rotatebox{90}{Obstacles} & \rotatebox{90}{Poles} & \rotatebox{90}{mIoU (\%)} \\ 

\hline\hline

SqueezeSeg~\cite{wu2018squeezeseg}
& 9.7 & 0.0 & 0.0 & 15.8 & 0.0 & 0.7 & 64.4 & 0.0 & 0.4 & 0.0 & 2.2 & 15.6 & 0.5 & 15.9 & 0.0 & 0.0 & 0.0 & 0.0 & 0.3 & 8.9 \\

SqueezeSegV2~\cite{wu2019squeezesegv2}
& 15.4 & 0.2 & 8.6 & 63.8 & 0.0 & 16.8 & 61.7 & 0.6 & 0.1 & 0.0 & 14.8 & 24.7 & 12.7 & 33.2 & 0.0 & 5.8 & 0.0 & 0.2 & 5.2 & 16.4  \\

DarkNet53~\cite{behley2019semantickitti}
& 15.2 & 0.8 & 6.1 & 68.5 & 0.0 & 15.5 & 63.8 & 0.4 & 0.3 & 0.0 & 17.3 & 23.8 & 13.3 & 35.6 & 0.0 & 6.3 & 0.0 & 3.9 & 7.6 & 17.2 \\

PolarNet~\cite{zhang2020polarnet}
& 23.8 & 10.1 & 18.2 & 69.7 & 9.6 & 49.1 & 58.5 & 0.0 & 11.3 & 0.0 & 28.3 & 37.6 & 24.8 & 42.8 & 0.0 & 14.8 & 0.0 & 8.0 & 11.0 & 23.9 \\

Cylinder3D~\cite{zhu2021cylindricalTPAMI} 
& 32.6 & 6.0 & 14.9 & 74.7 & 20.9 & 51.2 & 65.9 & 3.6 & 6.8 & 0.0 & 33.2 & 42.8 & 24.1 & 40.7 & 0.1 & 16.7 & 0.0 & 10.8 & 5.4 & 24.2\\
2DPASS~\cite{yan20222dpass} 
& 26.2 & 1.1 & 16.9 & 78.7 & 12.4 & 39.0 & 66.2 & 2.8 & 4.8 & 0.0 & 19.9 & 20.0 & 18.8 & 31.4 & 0.0 & 6.0 & 0.0 & 9.9 & 3.4 & 18.9 \\

PMF~\cite{zhuang2021perception} 
& 84.5 & 36.0 & 47.5 & 89.4 & 38.1 & 75.0 & 82.8 & 4.6 & 64.9 & 0.0 & 54.8 & 72.8 & 41.2 & 77.1 & 0.0 & 27.1 & 0.0 & 30.8 & 29.6 & 41.8 \\
\hline
Uni-modal baseline
& 42.3 & 5.6 & 13.7 & 75.2 & 0.6 & 32.2 & 65.6 & 0.2 & 13.3 & 0.0 & 30.6 & 50.5 & 12.3 & 53.9 & 0.1 & 6.9 & 0.0 & 6.1 & 9.2 & 21.8 \\
EPMF(Ours)
& 79.4 & 42.3 & 49.4 & 90.7 & 79.8 & 78.4 & 84.4 & 11.8 & 69.1 & 0.0 & 56.8 & 73.9 & 47.6 & 78.3 & 0.0 & 26.0 & 0.0 & 35.5 & 33.0 & \textbf{45.0} \\

\hline
\end{tabular}
}
\label{tab:a2d2_results_part1}
\end{table*}

\begin{table*}[t]
\centering
\caption{Comparisons on A2D2 test set - Part2. The \textbf{bold} numbers indicate the best results.}
  \scalebox{0.88}{
\begin{tabular}{l|ccccccccccccccccccc|c}
\hline
Method & \rotatebox{90}{RD restricted area} & \rotatebox{90}{Animals} & \rotatebox{90}{Grid structure} & \rotatebox{90}{Signal corpus} & \rotatebox{90}{Drivable cobbleston} & \rotatebox{90}{Electronic traffic} & \rotatebox{90}{Slow drive area} & \rotatebox{90}{Nature object} & \rotatebox{90}{Parking area} & \rotatebox{90}{Sidewalk} & \rotatebox{90}{Ego car} & \rotatebox{90}{Painted driv instr} & \rotatebox{90}{Traffic guide obj} & \rotatebox{90}{Dashed line} & \rotatebox{90}{RD normal street} & \rotatebox{90}{Sky} & \rotatebox{90}{Buildings} & \rotatebox{90}{Blurred area} & \rotatebox{90}{Rain dirt} & \rotatebox{90}{mIoU (\%)} \\ 

\hline\hline

SqueezeSeg~\cite{wu2018squeezeseg}
& 0.0 & 0.0 & 0.0 & 0.0 & 0.0 & 0.0 & 0.0 & 64.5 & 0.0 & 13.7 & 0.0 & 0.0 & 0.1 & 0.2 & 77.7 & 10.4 & 27.7 & 0.0 & 0.0 & 8.9 \\

SqueezeSegV2~\cite{wu2019squeezesegv2}
& 29.5 & 0.0 & 10.3 & 5.5 & 2.7 & 0.0 & 1.9 & 76.4 & 3.8 & 29.2 & 0.0 & 6.4 & 12.4 & 17.1 & 85.8 & 12.1 & 50.9 & 0.0 & 0.0 & 16.4  \\

DarkNet53~\cite{behley2019semantickitti}
& 38.7 & 0.0 & 10.8 & 4.4 & 3.3 & 0.0 & 0.0 & 77.9 & 3.1 & 31.5 & 0.0 & 9.4 & 7.3 & 15.7 & 86.4 & 12.9 & 55.2 & 0.0 & 0.0 & 17.2 \\

PolarNet~\cite{zhang2020polarnet}
& 55.6 & 0.0 & 14.8 & 11.9 & 7.0 & 0.0 & 4.4 & 81.6 & 12.8 & 42.5 & 0.0 & 12.7 & 11.5 & 31.8 & 90.3 & 9.2 & 57.0 & 0.0 & 0.0 & 23.9 \\

Cylinder3D~\cite{zhu2021cylindricalTPAMI}
& 57.7 & 0.0 & 20.5 & 10.6 & 10.4 & 1.5 & 2.5 & 82.9 & 9.1 & 47.4 & 0.0 & 19.1 & 8.0 & 33.8 & 89.7 & 13.0 & 63.5 & 0.0 & 0.0 & 24.2\\
2DPASS~\cite{yan20222dpass} 
& 37.0 & 0.0 & 13.8 & 6.7 & 5.0 & 0.0 & 0.3 & 74.8 & 5.9 & 35.1 & 0.0 & 8.6 & 10.0 & 11.9 & 85.6 & 6.9 & 57.5 & 0.0 & 0.0 & 18.9 \\
PMF~\cite{zhuang2021perception} 
& 65.4 & 0.0 & 46.6 & 22.6 & 26.3 & 0.0 & 3.7 & 91.7 & 17.8 & 48.6 & 0.0 & 48.9 & 64.1 & 69.2 & 94.1 & 53.5 & 78.4 & 0.0 & 0.0 & 41.8 \\
\hline
Uni-modal baseline 
& 45.7 & 0.0 & 14.0 & 2.7 & 9.1 & 0.0 & 0.1 & 74.6 & 3.1 & 38.1 & 0.0 & 12.3 & 11.5 & 35.5 & 90.2 & 14.2 & 59.0 & 0.0 & 0.0 & 21.8 \\
EPMF(Ours)
& 70.3 & 0.0 & 52.1 & 33.3 & 29.4 & 0.0 & 1.5 & 92.4 & 15.3 & 57.4 & 0.0 & 55.0 & 66.3 & 70.0 & 94.5 & 54.2 & 80.5 & 0.0 & 0.0 & \textbf{45.0} \\

\hline
\end{tabular}
}
\label{tab:a2d2_results_part2}
\end{table*}

\subsubsection{Evaluation Metrics}
To evaluate the performance of our method, we use the mean Intersection over Union (mIoU) as the evaluation metric following the official guidance in~\cite{behley2019semantickitti,caesar2020nuscenes}. To class $s$, the respective intersection over union $\rm{IoU}_s$ is defined by
\begin{equation}
    \rm{IoU}_s=\frac{|\mP_s\cap\mG_s|}{|\mP_s\cup\mG_s|},
\end{equation}
where $\mP_s$ is the set of point with a class prediction $s$, $\mG_s$ is the set of label for class $s$, and $|\cdot|$ represents the cardinality of the set. Then, the mIoU is formulated as $\rm{mIoU}=\frac{1}{S}\sum^S_{s=1}\rm{IoU}_s$.

\subsection{Implementation details}
\label{sec:implementation}
We implement the proposed method in PyTorch~\cite{paszke2019pytorch}, and use ResNet-34~\cite{he2016deep} and SalsaNext~\cite{cortinhal2020salsanext} as the backbones of the camera stream and LiDAR stream, respectively. Because we process the point clouds in the camera coordinates, we incorporate ASPP~\cite{chen2017deeplab} into the LiDAR stream network to adjust the receptive field adaptively. To leverage the benefits of existing image classification models, we initialize the parameters of ResNet-34 with the pre-trained ImageNet models from~\cite{paszke2019pytorch}. We also adopt hybrid optimization methods~\cite{zhang2020exploring} to train the networks~\wrt~different modalities,~\ie, SGD with Nesterov~\cite{nesterov1983method} for the camera stream and Adam~\cite{kingma2014adam} for the LiDAR stream. We train the networks for 50 epochs with a batch size of 8 on SemanticKITTI. On nuScenes and A2D2, the batch size is 24 with 150 epochs for training as the point clouds of these data sets are more sparse. The learning rate starts at 0.001 and decays to 0 with a cosine policy~\cite{loshchilov2016sgdr}. We tune the hyper-parameters of the objective function in Eq.~(\ref{eq:tsnet_obj}) using the weighting strategy proposed in~\cite{kendall2018multi}. We set the confidence threshold $\tau$ to 0.7 following ~\cite{zhuang2021perception}.\footnote{The ablation study of $\tau$ is in the supplementary materials of ~\cite{zhuang2021perception}.}To prevent over-fitting, a series of data augmentation strategies are used, including random horizontal flipping, random scaling, color jitter, 2D random rotation, and random cropping.

\subsection{Comparisons on benchmark data sets}
\label{sec:main_results}
\subsubsection{Results on SemanticKITTI-FV}
To evaluate our method on SemanticKITTI-FV, we compare EPMF with several state-of-the-art LiDAR-only methods including SalsaNext~\cite{cortinhal2020salsanext}, Cylinder3D~\cite{zhu2021cylindrical}, 2DPASS~\cite{yan20222dpass}, etc.  Following~\cite{cortinhal2020salsanext,jaritz2020xmuda,zhu2021cylindrical}, we use sequence 08 for validation. The remaining sequences (00-07 and 09-10) are used as the training set. 
We evaluate the release models of the state-of-the-art LiDAR-only methods on our data set. Because SPVNAS~\cite{tang2020searching} did not release its best model, we report the result of the best-released model (with 65G MACs). In addition, we re-implement two fusion-based methods,~\ie, RGBAL~\cite{Madawy2019RGBAL} and PointPainting~\cite{vora2020pointpainting} on our data set. To further understand the effectiveness of our proposed fusion strategies, we construct a \textit{uni-modal baseline}, which uses the same network architecture of LiDAR stream of the proposed two-stream network. Note that we do not use test time augmentation (TTA) during evaluation as this technique is time-consuming and cannot be adopted to real applications on the car. 

From Table~\ref{tab:semanti_kitti_results}, EPMF outperfoms the uni-modal baseline by 7.3\% in mIoU. Compared with PMF, our EPMF also achieves 2.0\% improvements in mIoU. However, EPMF performs slightly worse than the pre-trained model of 2DPASS~\cite{yan20222dpass} on SemanticKITTI-FV. Note that our EPMF is trained on SemanticKITTI-FV with only 16.03\% points of SemanticKITTI. For fair comparisons, we also train 2DPASS using the officially released code and evaluate the model on SemanticKITTI-FV. In this case, our EPMF outperforms 2DPASS by 4.1\% in mIoU.

\subsubsection{Results on nuScenes}
Following~\cite{zhu2021cylindrical}, to evaluate our method on more complex scenes, we compare EPMF with the state-of-the-art methods on the nuScenes LiDAR-seg validation set. The experimental results are shown in Table~\ref{tab:nuScenes_results}. Note that the point clouds of nuScenes are sparser than those of SemanticKITTI (35k points/frame vs. 125k points/frame). Thus, it is more challenging for 3D segmentation tasks. In this case, EPMF achieves the best performance on nuScenes validation set. Specifically, EPMF outperforms the best LiDAR-only method, ~\ie, SphereFormer, by 2.2\% in mIoU. Compared with PMF, our EPMF achieves 3.7\% improvements in mIoU. 

In addition, we further provide the results on nuScenes test set. Different from the results on the nuScenes validation set, existing methods always use additional techniques, including test-time augmentation or fine-tuning with class re-sampling that improves performance on rare classes, to pursue better results on the leaderboard. Unlike existing methods, we directly evaluate our model on the test set without additional techniques. For fair comparisons, we also evaluate the performance of the released models of SphereFormer without TTA or fine-tuning. As shown in Table~\ref{tab:nus_test_set}, our EPMF outperforms SphereFormer~\cite{lai2023spherical} by 1.1\% in mIoU under the same evaluation settings. Compared with PMF, our EPMF consistently achieves 3.7\% improvements in mIoU on nuScenes test set.
These results are consistent with our expectations. Since EPMF incorporates RGB images, our fusion strategy is capable of addressing such challenging segmentation under extremely sparse point clouds.

\subsubsection{Results on A2D2}
To further evaluate the performance of our method on more sparse and irregular point clouds, we conduct experiments on A2D2 and compare our EPMF to existing methods. For a fair comparison, we select the model with the best validation performance and report the evaluation results on the test set. Note that both Cylinder3D and 2DPASS did not report the results on A2D2 test set, we also train Cylinder3D and 2DPASS using the officially released code and report the results on the test set. As shown in Table~\ref{tab:a2d2_results_part1} and Table~\ref{tab:a2d2_results_part2}, both our PMF and EPMF outperform the LiDAR-only method by a large margin. Specifically, our EPMF outperforms Cylinder3D by 20.8\% in mIoU. Compared to uni-modal baseline, EPMF achieves 23.2\% improvements in mIoU, which indicates the effectiveness of our proposed fusion strategies. Moreover, our EPMF outperforms PMF consistently, with 3.2\% improvement in mIoU.

\begin{figure*}
\centering
    \subfigure[Comparisons on SemanticKITTI-FV]{
        \includegraphics[width=0.31\linewidth]{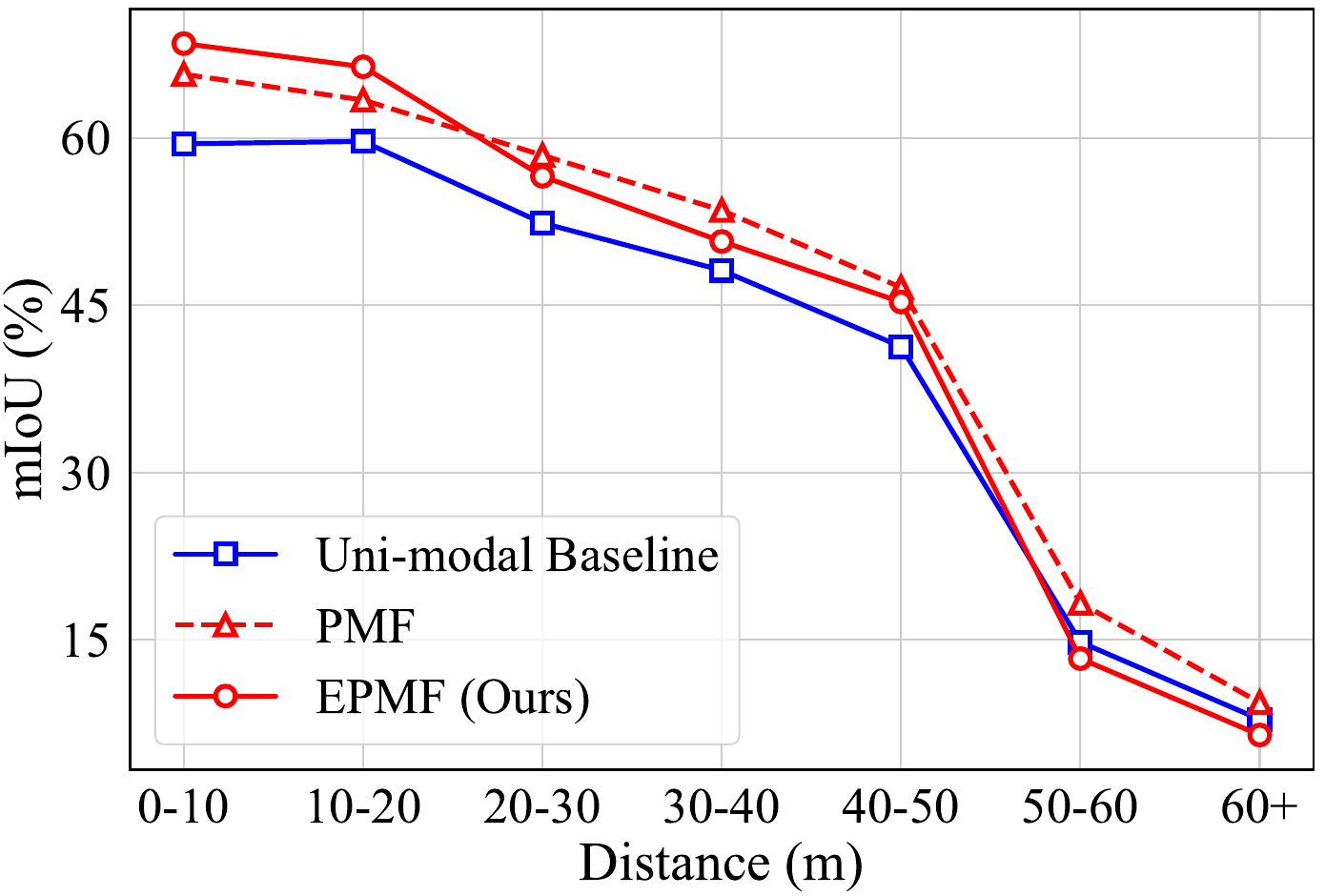}
    }
    \subfigure[Comparisons on nuScenes]{
        \includegraphics[width=0.31\linewidth]{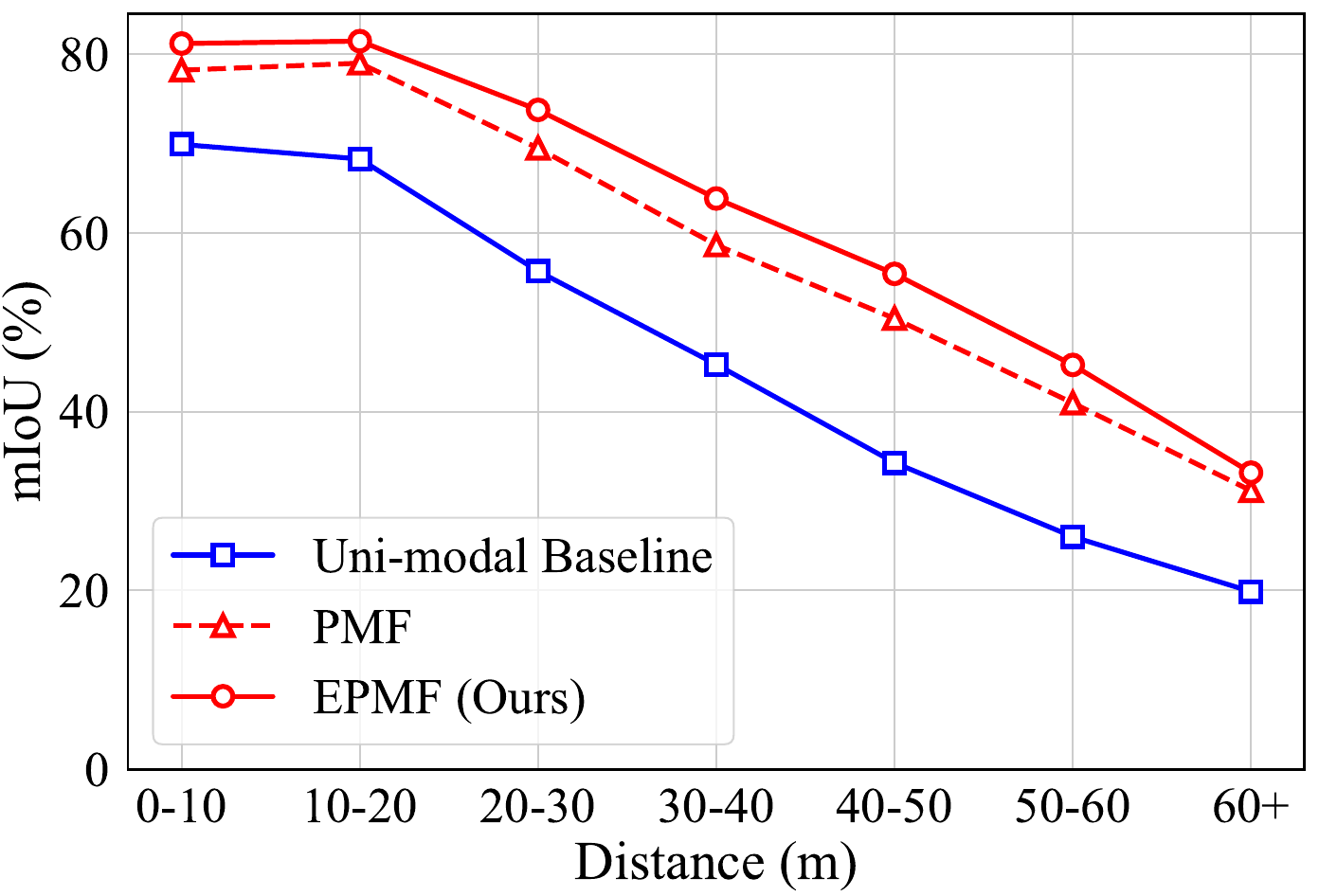}
    }
    \subfigure[Comparisons on A2D2]{
        \includegraphics[width=0.31\linewidth]{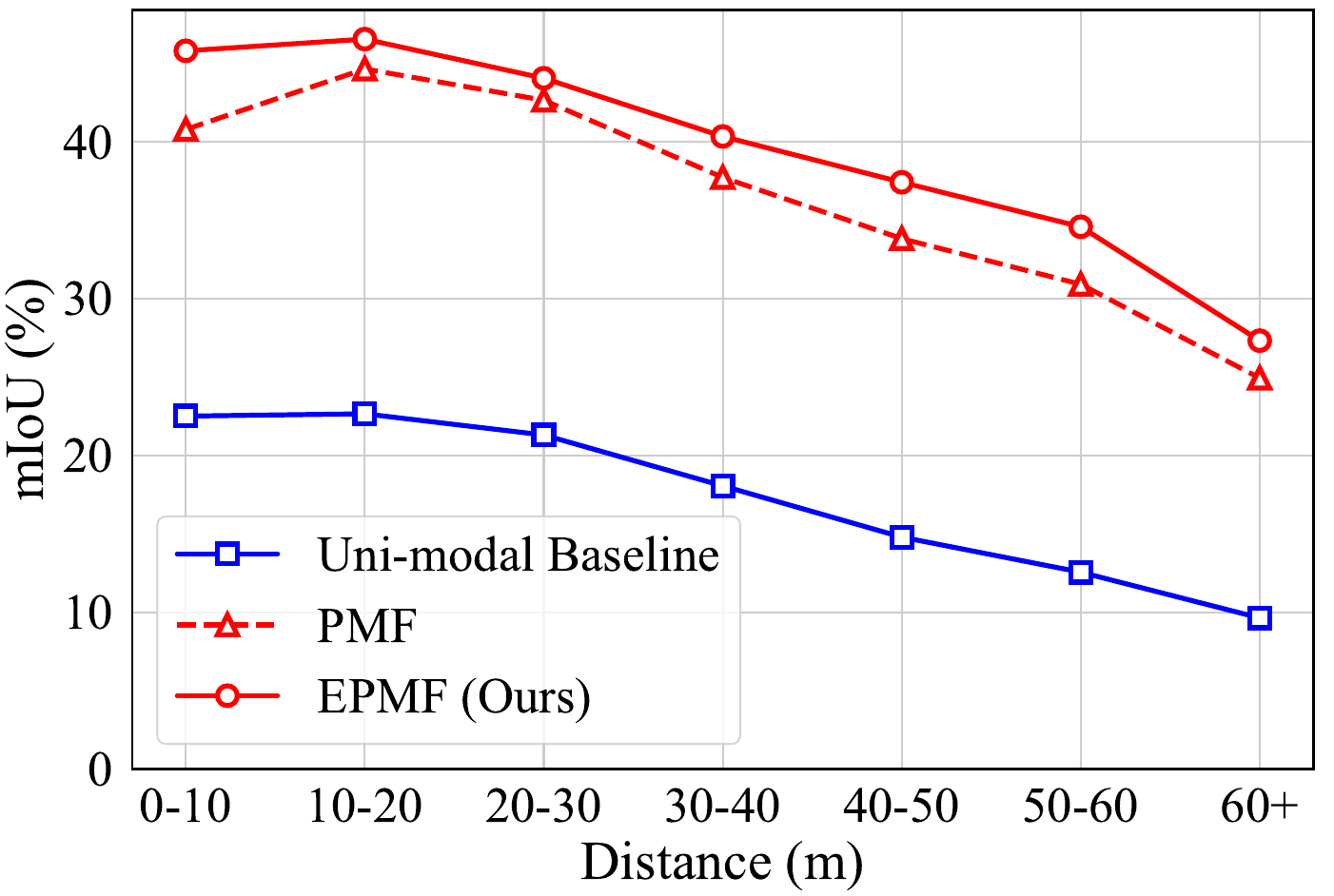}
    }
     \caption{Distance-based evaluation on SemanticKITTI-FV, nuScenes and A2D2. As the distance increases, the point cloud becomes sparser.}
    
\label{fig:cmp_distance_perf}
\end{figure*}

\begin{table}
    \centering
    \caption{Comparisons of the number of classes at different distances.}
    \scalebox{0.88}{
    \begin{tabular}{c|ccccccc}
    \hline
        \multirow{2}{*}{Dataset} & \multicolumn{7}{c}{Distance (m)} \\ 
        \cline{2-8}
         & 0-10 & 10-20 & 20-30 & 30-40 & 40-50 & 50-60 & 60+ \\
         \hline \hline
         SemanticKITTI-FV & 18 & 19 & 19 & 19 & 19 & 11 & 8 \\
         nuScenes & 16 & 16 & 16 & 16 & 16 & 16 & 16\\
         A2D2 & 33 & 35 & 35 & 34 & 34 & 33 & 34 \\ \hline
    \end{tabular}
    }
    
    \label{tab:cmp_benchmark_datasets_cls}
\end{table}
\begin{table}
    \centering
    \caption{Percentage of point cloud distribution (\%) at different distances.}
    \scalebox{0.88}{
    \begin{tabular}{c|ccccccc}
    \hline
        \multirow{2}{*}{Dataset} & \multicolumn{7}{c}{Distance (m)} \\ 
        \cline{2-8}
          & 0-10 & 10-20 & 20-30 & 30-40 & 40-50 & 50-60 & 60+ \\
         \hline \hline
         SemanticKITTI-FV & 34.7 & 41.5 & 12.9 & 5.2 & 2.7 & 1.5 & 1.5 \\
         nuScenes & 70.8 & 15.7 & 6.3 & 3.2 & 1.8 & 1.0 & 1.2 \\
         A2D2 & 27.4 & 36.9 & 19.5 & 9.4 & 4.1 & 1.6 & 1.1\\ 
         \hline
    \end{tabular}
    }
    
    \label{tab:cmp_benchmark_datasets_points}
\end{table}

\subsection{Distance-based Evaluation}
\label{sec:distance_eval}
In 3D LiDAR perception, the point cloud becomes sparser with the increase of perception distance. As long-range perception is important to the safety of autonomous cars, we further conduct a distance evaluation on the benchmark datasets and investigate the performance of our method under different distances. As shown in Figure~\ref{fig:cmp_distance_perf}, both PMF and EPMF outperform the uni-modal baseline by a large margin under different distances on nuScenes and A2D2, which indicates that our fusion strategy can incorporate the information from RGB images effectively. 

We also notice that EPMF cannot consistently outperform PMF or the uni-modal baseline on SemanticKITTI-FV. We argue that this phenomenon is mainly caused by the following reasons. First, we compare the number of classes at different distances~\wrt the validation set of SemanticKITTI-FV and nuScenes, as well as the test set of A2D2. From Table~\ref{tab:cmp_benchmark_datasets_cls}, the number of classes on SemanticKITTI-FV decreases at long distances. Since it is difficult for our method to beat the baseline on all the semantic classes of SemanticKITTI-FV, EPMF performs worse than the uni-modal baseline when the distance is larger than 50m on SemanticKITTI-FV, which covers only 3\% of point clouds (See Table~\ref{tab:cmp_benchmark_datasets_points}). Second, to make a trade-off between efficiency and effectiveness, we insert a down-sampling operation into the contextual module of LiDAR stream to improve its efficiency. As the point clouds of SemanticKITTI are dense, reducing the resolution of point cloud features also leads to higher classification errors at long distances. Nevertheless, EPMF achieves better performance when the distance is less than 20m which covers 76.2\% of the points. Overall, our EPMF outperforms PMF by 2.0\% in mIoU with 2.06$\times$ acceleration on SemanticKITTI-FV.

\begin{table}
\centering
\caption{Inference time of different methods on SemanticKITTI using TensorRT. "-" indicates the results that are not available. For a fair comparison, Cylinder3D is accelerated by sparse convolution.}

\scalebox{1.0}
{
\begin{tabular}{l|cccccc}
    \hline
    Method & \#FLOPs & \#Params. & Inference time & mIoU\\
    \hline\hline 
    PointPainting~\cite{vora2020pointpainting} & 51.0 G & 28.1 M & 2.3 ms & 54.5\% \\
    RGBAL~\cite{Madawy2019RGBAL} & 55.0 G & 13.2 M & 2.7 ms & 56.2\% \\
    SalsaNext~\cite{cortinhal2020salsanext} & 31.4 G & 6.7 M & 1.6 ms & 59.4\% \\
    Cylinder3D~\cite{zhu2021cylindrical} &  - & 55.9 M & 62.5 ms & 64.9\%  \\
    PMF~\cite{zhuang2021perception} & 854.7 G & 36.3 M & 22.3 ms & 63.9\% \\ 
    \hline
    EPMF (Ours) & 418.0 G & 34.2 M & 14.2 ms & \textbf{65.9\%} \\ 
    \hline
\end{tabular}
}
\label{tab:flops_pmf}
\end{table}

\begin{figure*}[t]
\centering
    \includegraphics[width=\linewidth]{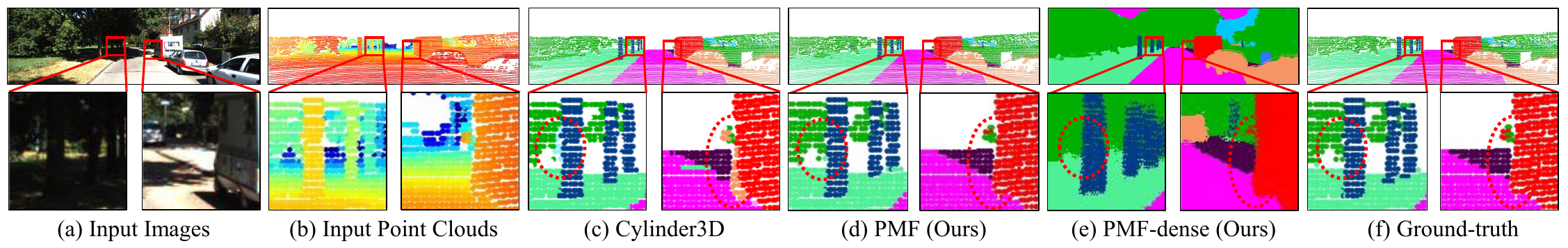}
   \caption{Qualitative results on SemanticKITTI-FV. The red dashed circle indicates the difference between the results of PMF and the baseline.}
\label{fig:quality_results}
\end{figure*}

\subsection{Efficiency analysis}
\label{sec:efficiency}
In this section, we evaluate the efficiency of EPMF on GeForce RTX 3090. Note that we consider the efficiency of PMF in two aspects. First, since predictions of the camera stream are fused into the LiDAR stream, we remove the decoder of the camera stream to speed up the inference. Second, both our PMF and EPMF are built on 2D convolutions and can be easily optimized by existing inference toolkits,~\eg, TensorRT. In contrast, Cylinder3D is built on 3D sparse convolutions~\cite{graham20183d} and is difficult to be accelerated by TensorRT. We report the inference time of different models optimized by TensorRT in Table~\ref{tab:flops_pmf}. From the results, our PMF achieves the best performance on nuScenes and is $2.8\times$ faster than Cylinder3D (22.3 ms vs. 62.5 ms) with fewer parameters. While compared to PMF, our EPMF achieves $1.6\times$ acceleration (14.2 ms vs. 22.3 ms) with 2.0\% improvements in mIoU.
\begin{figure}
   \includegraphics[width=\linewidth]{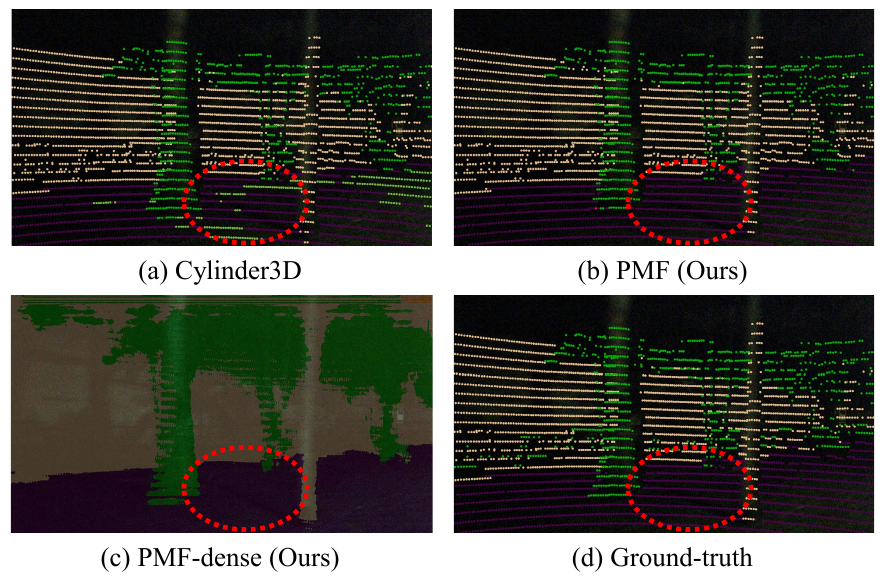}
   \caption{Qualitative results on nuScenes. We use the corresponding images (night) as the background of both the predictions and labels. We highlight the difference between the results of PMF and the baseline with the red dashed circle.}
\label{fig:quality_results_nus}
\end{figure}

\begin{figure*}
    \centering
        \includegraphics[width=\linewidth]{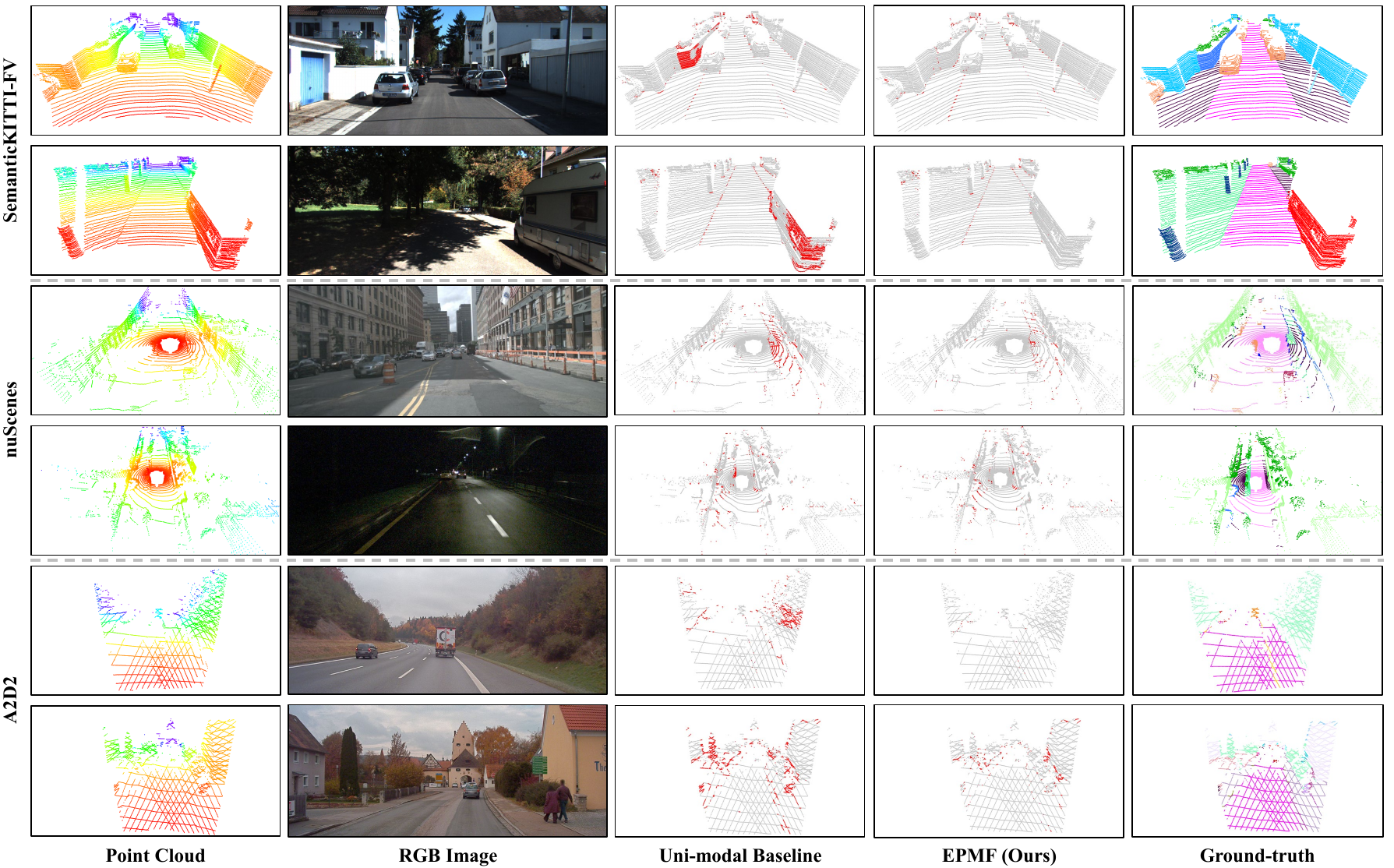}
        \caption{Qualitative results on three benchmark data sets. We show the error maps of both our method and the uni-modal baseline, in which the red points indicate mispredictions. Zoom in for more details.}
        \label{fig:error_map}
\end{figure*}

\subsection{Qualitative evaluation}
\label{sec:qualitative_eval}
To better understand the benefits of PMF, we visualize the predictions of PMF on the benchmark data sets. From Figure~\ref{fig:quality_results}, compared with Cylinder3D, PMF achieves better performance at the boundary of objects. For example, as shown in Figure~\ref{fig:quality_results} (d), the truck segmented by PMF has a more complete shape. More critically, PMF is robust in different lighting conditions. Specifically, as illustrated in Figure~\ref{fig:quality_results_nus}, PMF outperforms the baselines on more challenging scenes (\eg, night). In addition, as demonstrated in Figure~\ref{fig:quality_results} (e) and Figure~\ref{fig:quality_results_nus} (c), PMF generates dense segmentation results that combine the benefits of both the camera and LiDAR, which are significantly different from existing LiDAR-only and fusion-based methods.

In Figure~\ref{fig:error_map}, we show the error maps of both EPMF and the uni-modal baseline on three benchmark data sets under different scenes. From the results, our EPMF can fully utilize both information from point clouds and RGB images and thus performs well in some challenging scenes with extremely sparse point clouds or poor lighting conditions.

\begin{figure}
\centering
\includegraphics[width=0.98\linewidth]{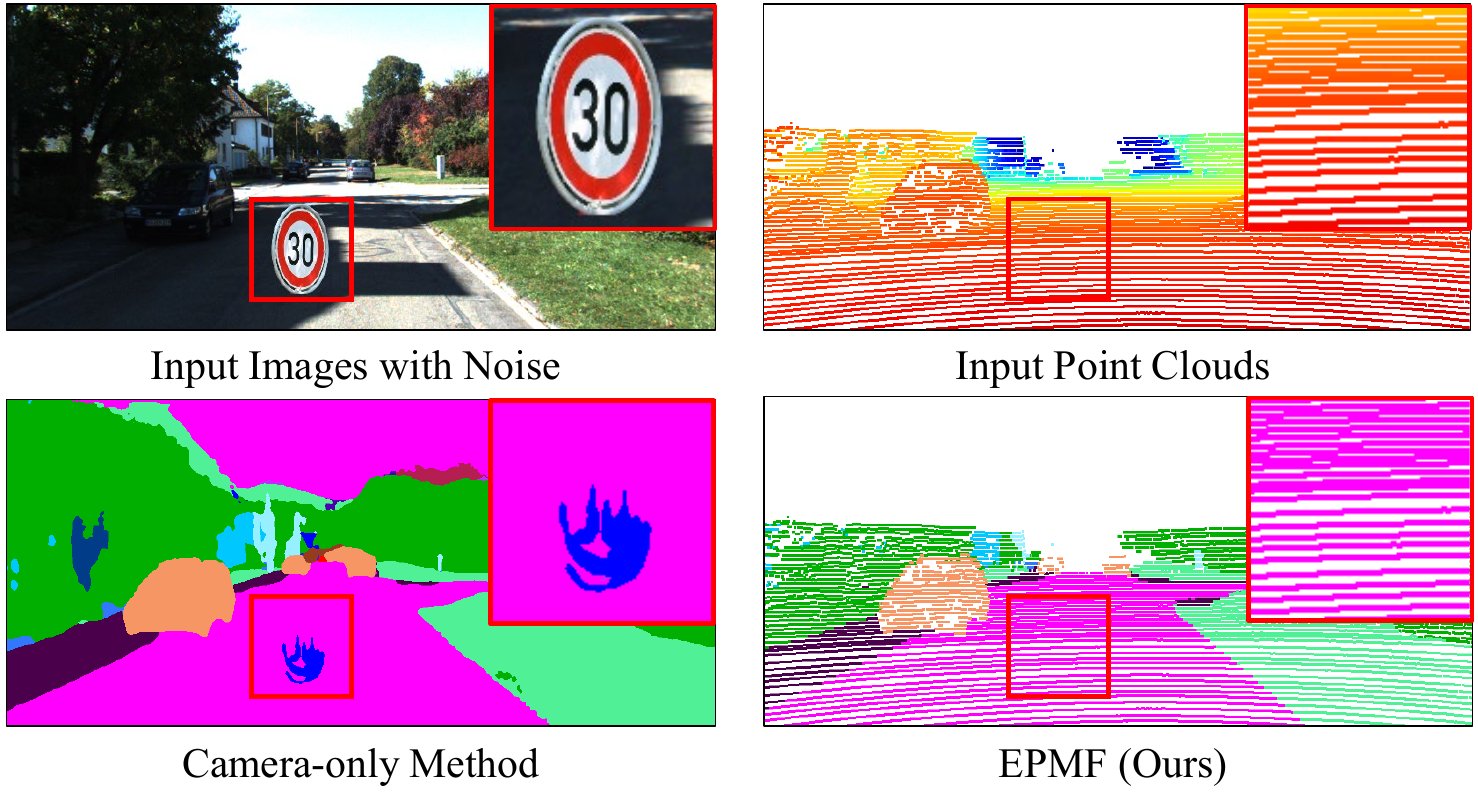}
\caption{Comparisons of EPMF and camera-only methods on adversarial samples. The camera-only methods use only RGB images as inputs, while PMF uses both images and point clouds as inputs. We highlight the inserted traffic sign with a red box.}
\label{fig:visual_anti_attack}
\end{figure}

\subsection{Robustness analysis}
\label{sec:abla_anti_attack}
To investigate the robustness of EPMF on adversarial samples, we first insert extra objects (\eg, a traffic sign) into the images and keep the point clouds unchanged. In addition, we implement a camera-only method,~\ie, FCN~\cite{Long2015FullyCN}, on SemanticKITTI as the baseline. Note that we do not use any adversarial training technique during training. As demonstrated in Figure~\ref{fig:visual_anti_attack}, the camera-only methods are easily affected by changes in the input images. In contrast, because EPMF integrates reliable point cloud information,
the noise in the images is reduced during feature fusion and imposes only a slight effect on the model performance.

To investigate the performances of EPMF under different lighting conditions, we evaluate the performance of EPMF on nuScenes validation set at day/night time. We initialize SegFormer-B5 by the officially released weights on CityScape and train the model SegFormer-B5 with the camera or LiDAR as inputs. Note that some of the classes may not appear at night, we only report the mIoU of the available classes. From Table~\ref{tab:cmp_segformer_night}, as the camera only provides limited information at night, SegFormer with camera-only inputs performs worse than the LiDAR-only counterpart. By fusing both information from the camera and LiDAR, EPMF achieves better performance at night, with 15.1\% improvements in mIoU compared to SegFormer-B5.


\begin{table}
    \centering
    \caption{Comparisons to SegFormer on nuScenes at day/night time. \textbf{C} and \textbf{L} indicate the model with camera or LiDAR as input, respectively. \textbf{L+C} indicates the model with both camera and LiDAR inputs.}
    \scalebox{1.0}{
    \begin{tabular}{c|c|cc}
        \hline
         Method & Input & Day & Night \\
         \hline\hline
         SegFormer-B5~\cite{xie2021segformer} & C & 67.6\% & 41.6\% \\
         SegFormer-B5~\cite{xie2021segformer} & L & 56.6\% & 46.9\% \\
         \hline
         EPMF (Ours) & L+C & 81.4\% & 62.0\% \\
         \hline
    \end{tabular}
    }
    \label{tab:cmp_segformer_night}
\end{table}

\begin{table}
\centering
\caption{Ablation study for the network components on the SemanticKITTI-FV validation set. \textbf{PP} denotes perspective projection. \textbf{RF} denotes the residual-based fusion module. \textbf{PL} denotes perception-aware loss. The \textbf{bold} number is the best result.}

  \scalebox{1.0}{
    \begin{tabular}{cccccc|c}
    \hline
    &Baseline & PP & ASPP & RF &  PL & mIoU (\%) \\
    \hline\hline
    1&\checkmark & & & & & 57.2 \\
    \hline
    2&\checkmark & \checkmark & & & & 57.6 \\
    3&\checkmark & \checkmark & \checkmark & & & 59.7 \\
    4&\checkmark &  & \checkmark & \checkmark & & 55.8 \\
    5&\checkmark & \checkmark & \checkmark & \checkmark & & 61.7 \\
    6&\checkmark & \checkmark & \checkmark & \checkmark & \checkmark & \textbf{63.9} \\
    \hline
    \end{tabular}
    }

\label{tab:ablation_components}
\end{table}

\section{Ablation Study}

\subsection{Effect of network components}
We study the effect of the network components of PMF,~\ie, perspective projection, ASPP, residual-based fusion modules, and perception-aware loss. The experimental results are shown in Table~\ref{tab:ablation_components}. Since we use only the front-view point clouds of SemanticKITTI, we train SalsaNext as the baseline on our data set using the officially released code. Comparing the first and second lines in Table~\ref{tab:ablation_components}, perspective projection achieves only a 0.4\% mIoU improvement over spherical projection with LiDAR-only input. In contrast, comparing the fourth and fifth lines, perspective projection brings a 5.9\% mIoU improvement over spherical projection with multimodal data inputs. From the third and fifth lines, our fusion modules bring 2.0\% mIoU improvement to the fusion network. Moreover, comparing the fifth and sixth lines, the perception-aware losses improve the performance of the network by 2.2\% in mIoU.

\begin{table}
    \centering
    \caption{Ablation study for the proposed improved techniques on SemanticKITTI-FV validation set.}
    \scalebox{1.0}{
    \begin{tabular}{l|ccc}
        \hline
         Proposed strategies & mIoU & \#FLOPs & \#Params. \\
         \hline\hline
         PMF~\cite{zhuang2021perception} & 63.9\% & 859.7 G & 36.4 M \\
         +dropping decoder of camera stream & 63.9\% & 854.7 G & 36.3 M \\ 
         \hdashline
         +cross-modal alignment and crop & 64.4\% & 739.9 G & 36.3 M \\
         +improved contextual module & 64.8\% & 418.0 G & 34.2 M \\
         +fusing high-level LiDAR feature & 65.9\% & 418.0 G & 34.2 M \\
         \hline
    \end{tabular}
    }
    \label{tab:abla_tech}
\end{table}

\begin{table}[t]
\centering
\caption{Model performance with different image masked ratios on SemanticKITTI-FV. We report the mIoU (\%) with/without fine-tuning.}
\scalebox{0.99}{
\begin{tabular}{c|cccccc}
    \hline
     Masked Ratio (\%) & 0 & 10 & 20 & 30 & 40 & 50 \\
     \hline\hline
     without fine-tuning & 65.9 & 65.5 & 64.8 & 63.0 & 60.0 & 54.8 \\
     with fine-tuning & 65.7 & 65.4 & 65.0 & 64.2 & 62.9 & 60.6 \\
     \hline
\end{tabular}
}
\label{tab:impact_mask_rgb}
\end{table}

\subsection{Effect of perception-aware loss}
\begin{figure}
\centering
\includegraphics[width=0.98\linewidth]{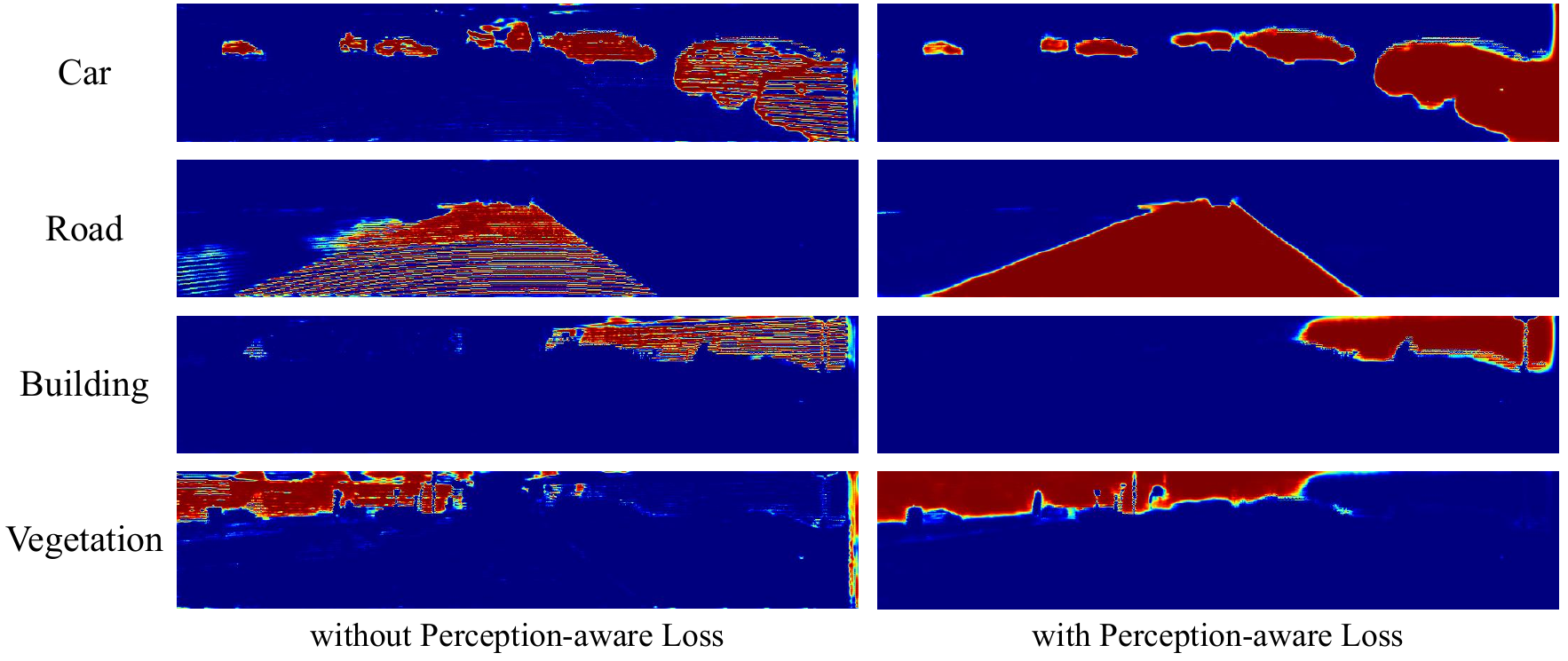}
\caption{Comparisons of the predictions~\wrt~the networks trained with and without perception-aware loss.}
\label{fig:visual_pred_conf}
\end{figure}

To investigate the effect of perception-aware loss, we visualize the predictions of the LiDAR stream networks with and without perception-aware loss in Figure~\ref{fig:visual_pred_conf}. From the results, perception-aware loss helps the LiDAR stream capture the perceptual information from the images. For example, the model trained with perception-aware loss learns the complete shape of cars, while the baseline model focuses only on the local features of points. As the perception-aware loss introduces the perceptual difference between the RGB images and the point clouds, it enables an effective fusion of the perceptual information from the data of both modalities. As a result, our PMF generates dense predictions that combine the benefits of both the images and point clouds.

\subsection{Effect of the improved techniques}
\label{sec:eff_improved}
In this section, we explore the effectiveness of the proposed improved techniques on SemanticKITTI-FV. As shown in Table~\ref{tab:abla_tech}, on top of PMF, dropping the decoder of the camera stream saves 5.0 GFLOPs in computation budgets without performance degradation. The proposed CAC further reduces 114.8 GFLOPs in computation costs by removing the useless area of RGB images. With the proposed improved contextual module, we achieve $2.04\times$ acceleration compared to PMF (418.0 GFLOPs vs. 854.7 GFLOPs) with 0.9\% improvement in mIoU. By fusing the high-level features of LiDAR stream into the camera stream, we further achieve 1.1\% improvements in mIoU.

When conducting CAC, if the LiDAR has a larger vertical FOV than cameras, keeping point clouds outside the image also results in partial blank image inputs. To investigate the impact of these areas without image information, we partially mask the RGB image from bottom to top with ratios from 10\% to 50\%, and evaluate the model performance on SemanticKITTI-FV. As shown in Table~\ref{tab:impact_mask_rgb}, masking 10\% of RGB image only results in 0.4\% degradation of mIoU. With the increasing of the masked ratio, the model performance suffers from significant degradation. Nevertheless, the impact of masked images can be mitigated by introducing the image mask into training or fine-tuning.

\section{Conclusion}
In this work, we have proposed a perception-aware multi-sensor fusion scheme for 3D LiDAR semantic segmentation. Unlike existing methods that conduct feature fusion in the LiDAR coordinate system, we project the point clouds to the camera coordinate system to enable a collaborative fusion of the perceptual features from the two modalities. By fusing complementary information from both cameras and LiDAR, PMF is robust in complex outdoor scenes with extremely sparse point clouds or poor lighting conditions. Moreover, we propose EPMF which improves the efficiency and effectiveness of PMF. Specifically, we introduce cross-modal alignment and cropping to reduce unnecessary computation. Besides, we also adjust the architecture of the fusion network by fusing high-level features into the camera stream and exploring the design of contextual modules under perspective projection. The experimental results on three benchmarks show the superiority of our method. In the future, we will extend EPMF to other challenging tasks,~\eg, object detection and semantic scene completion.

\vspace{-0.1in}
\ifCLASSOPTIONcompsoc
  \section*{Acknowledgments}
\else
  \section*{Acknowledgment}
\fi

This work was partially supported by
 National Natural Science Foundation of China (NSFC) 62072190, Key-Area Research and Development Program of Guangdong Province 2018B010107001, Guangdong Introducing Innovative and Enterpreneurial Teams 2017ZT07X183.

\ifCLASSOPTIONcaptionsoff
  \newpage
\fi
\bibliographystyle{abbrv}
{
	\bibliography{longstrings,reference}
}
\vspace{-0.15in}
\begin{IEEEbiography}[{\includegraphics[width=1in,height=1.25in,clip,keepaspectratio]{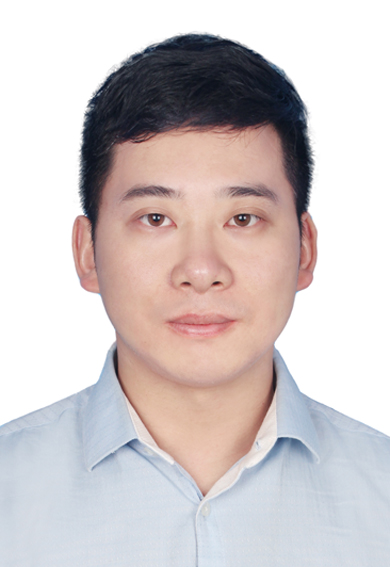}}]{Mingkui Tan}
is currently a Professor with the School of Software Engineering, South China University of Technology, Guangzhou, China. He received the Bachelor Degree in Environmental Science and Engineering in 2006 and the Master Degree in Control Science and Engineering in 2009, both from Hunan University in Changsha, China. He received the Ph.D. degree in Computer Science from Nanyang Technological University, Singapore, in 2014. From 2014-2016, he worked as a Senior Research Associate on computer vision in the School of Computer Science, University of Adelaide, Australia. His research interests include machine learning, sparse analysis, deep learning and large-scale optimization.
\end{IEEEbiography}

\begin{IEEEbiography}[{\includegraphics[width=1in,height=1.25in,clip,keepaspectratio]{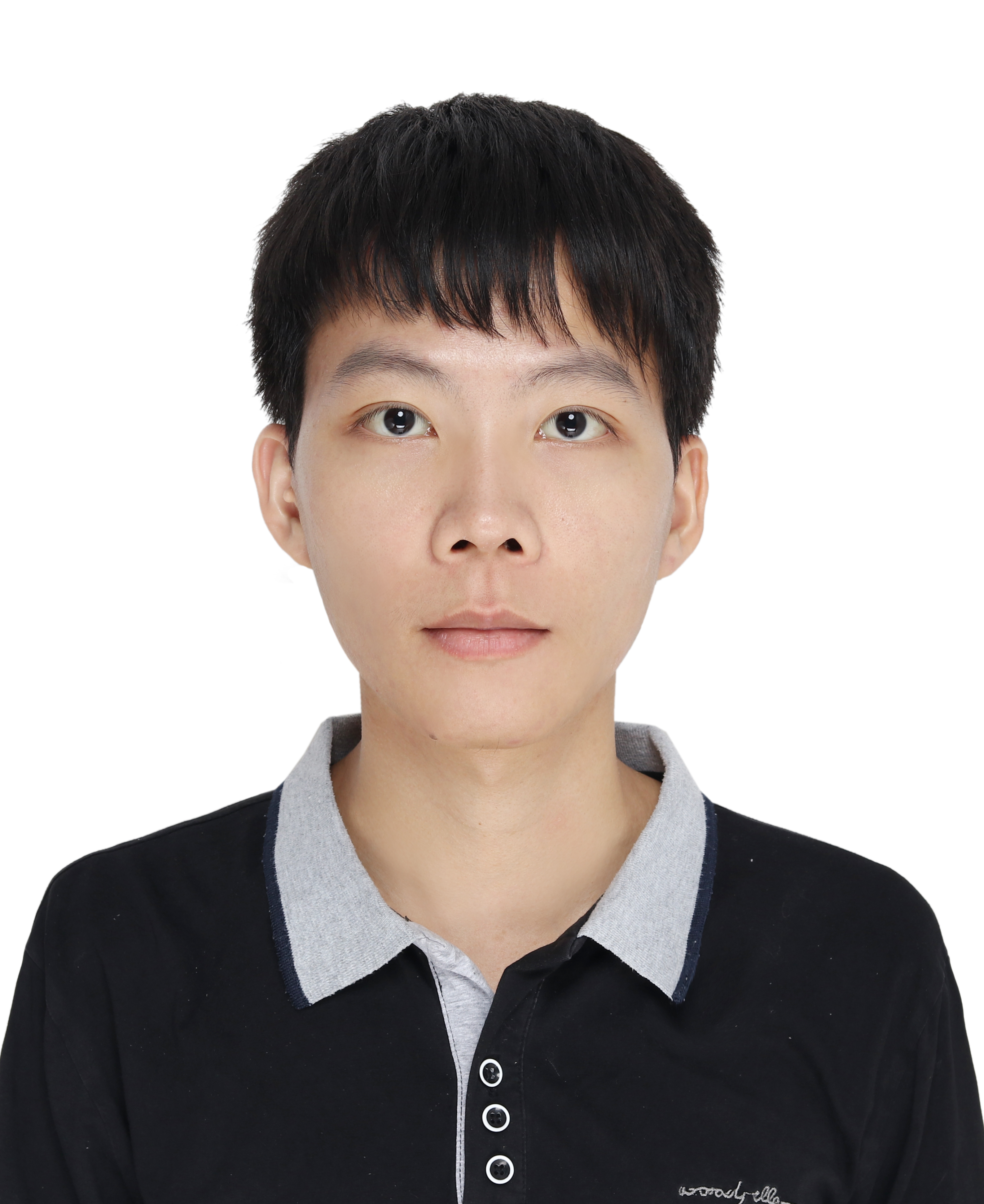}}]{Zhuangwei Zhuang}
is a Ph.D. student in the School of Software Engineering at South China University of Technology, and is currently working as an intern at RoboSense, Shenzhen, China. He received his Bachelor Degree in Automation and Engineering in 2016 and Master Degree in Software Engineering in 2018, both from South China University of Technology in Guangzhou, China. His research interests include model compression and 3D scene understanding for autonomous driving.
\end{IEEEbiography}
\begin{IEEEbiography}[{\includegraphics[width=1in,height=1.25in,clip,keepaspectratio]{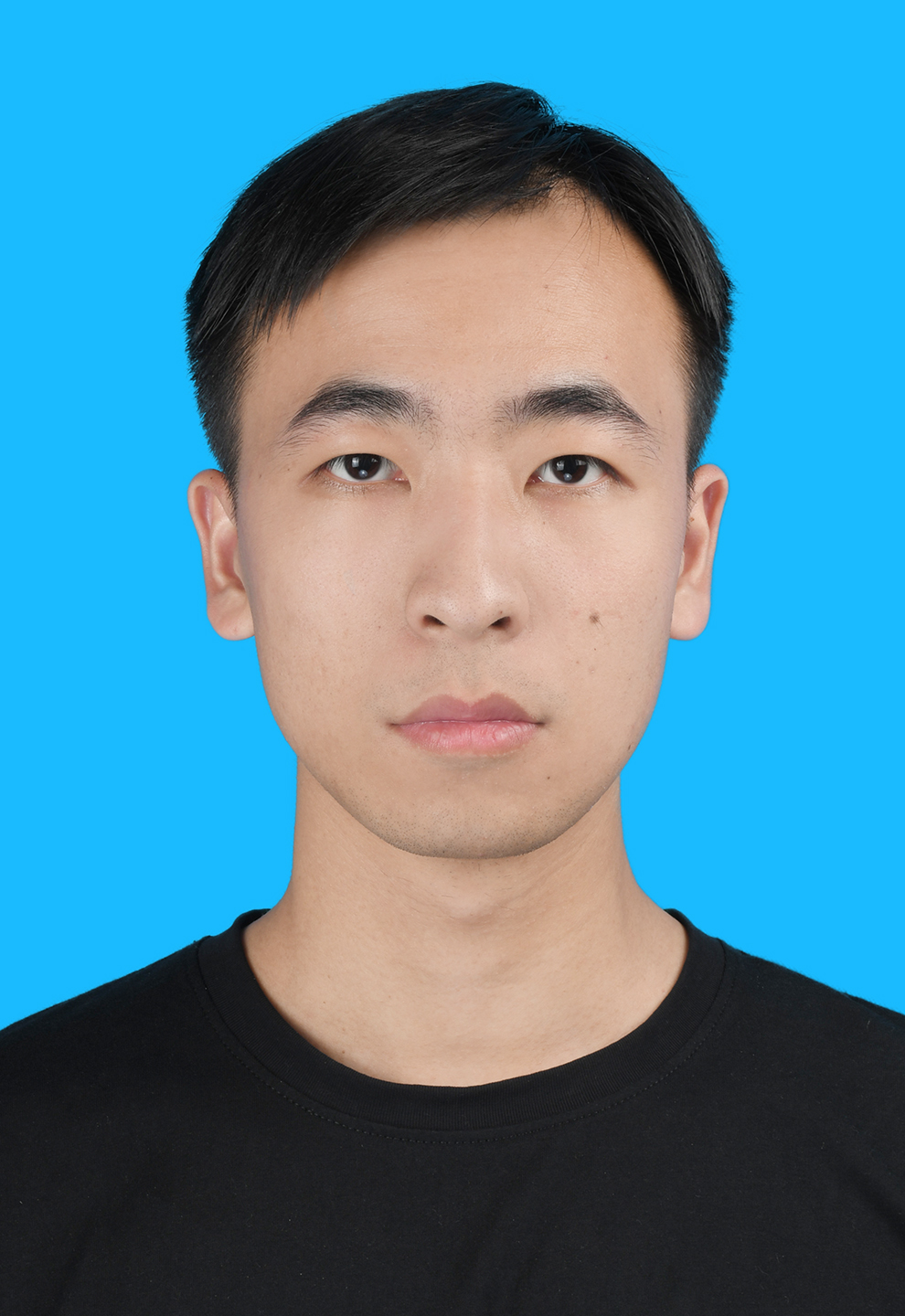}}]{Sitao Chen}
is a Master student in the School of Software Engineering at South China University of Technology.  He received his Bachelor Degree in the School of Mechanical \& Automotive in 2023 from South China University of Technology in Guangzhou, China. His research interests include 3D scene understanding for autonomous driving.
\end{IEEEbiography}
\begin{IEEEbiography}[{\includegraphics[width=1in,height=1.25in,clip,keepaspectratio]{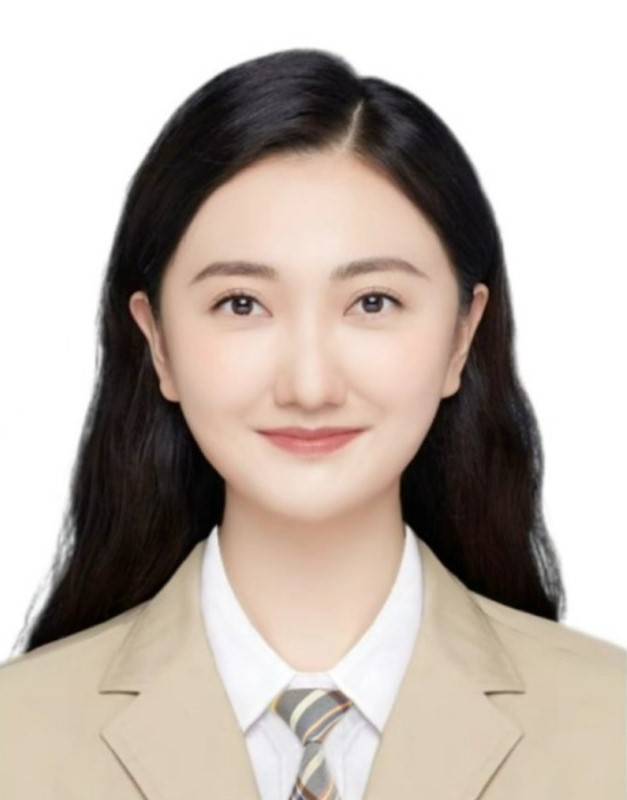}}]{Rong Li}
is currently an Associate Researcher at the Hong Kong University of Science and Technology (Guang Zhou). She received her Bachelor's Degree in 2019 and Master's Degree in 2022, both from the School of Software Engineering at South China University of Technology, China. Her research interests include LiDAR perception. 
\end{IEEEbiography}
\clearpage
\begin{IEEEbiography}[{\includegraphics[width=1in,height=1.25in,clip,keepaspectratio]{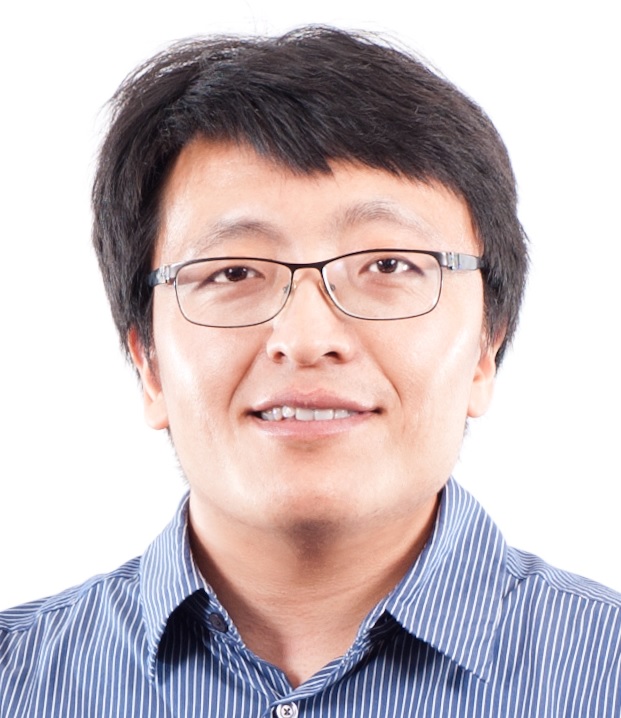}}]{Kui Jia}
is currently a Professor at the School of Electronic and Information Engineering, South China University of Technology, Guangzhou, China. He received his Bachelor Degree in marine engineering from Northwestern Polytechnic University, Xi’an, China, in 2001, the Master Degree in electrical and computer engineering from National University of Singapore in 2003, and the Ph.D. degree in computer science from Queen Mary University of London, U.K., in 2007. His recent research focuses on theoretical deep learning and its applications in vision and robotic problems, including deep learning of 3D data and deep transfer learning.
\end{IEEEbiography}
\begin{IEEEbiography}[{\includegraphics[width=1in,height=1.25in,clip,keepaspectratio]{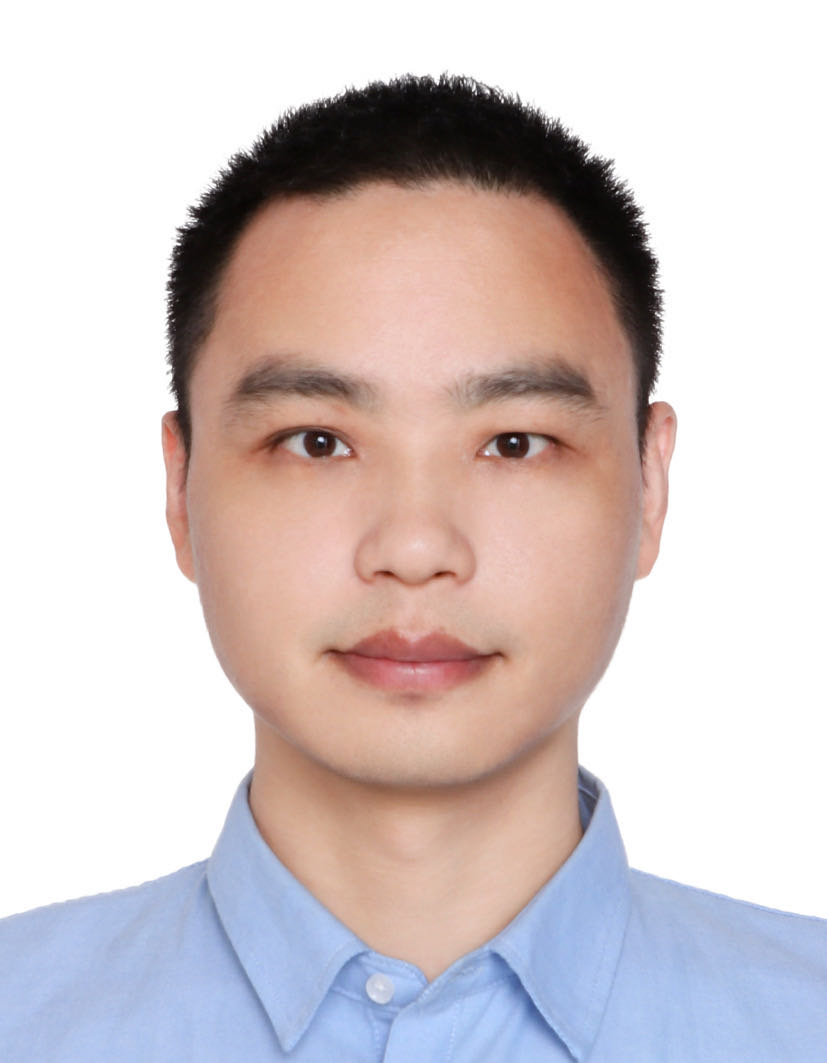}}]{Qicheng Wang}
is Ph.D. student in the Department of Mathematics at the Hong Kong University of Science and Technology. He received his Bachelor Degree in Electronic Engineering in 2007 from Tsinghua University, Beijing. His research interest lies in neural network and computer vision for autonomous driving.
\end{IEEEbiography}
\begin{IEEEbiography}[{\includegraphics[width=1in,height=1.25in,clip,keepaspectratio]{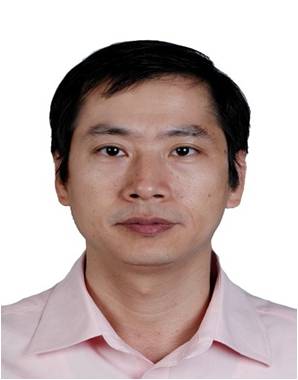}}]{Yuanqing Li}
is currently a Professor with the School of Automation and Engineering, South China University of Technology, Guangzhou, China. He received the Bachelor Degree in applied mathematics from Wuhan University, Wuhan, China, in 1988, the Master Degree in applied mathematics from South China Normal University, Guangzhou, China, in 1994, and the Ph.D. degree in control theory and applications from the South China University of Technology, Guangzhou, in 1997. His research interests include blind signal processing, sparse representation, machine learning, brain-computer interface, and EEG and functional magnetic resonance imaging data analysis.
\end{IEEEbiography}




\end{document}